\documentclass[a4paper,fleqn]{cas-dc}

\usepackage[numbers]{natbib}
\usepackage{enumitem}
\usepackage{graphicx}
\usepackage{multirow}
\usepackage{color}
\usepackage{bm}
\usepackage{hyperref} 

\def\tsc#1{\csdef{#1}{\textsc{\lowercase{#1}}\xspace}}
\tsc{WGM}
\tsc{QE}
\tsc{EP}
\tsc{PMS}
\tsc{BEC}
\tsc{DE}

\begin{document}
\let\WriteBookmarks\relax
\def\floatpagepagefraction{1}
\def\textpagefraction{.001}
\shorttitle{Fine-Grained Few Shot Learning with Foreground Object Transformation}
\shortauthors{Chaofei Wang et~al.}

\title [mode = title]{Fine-Grained Few Shot Learning with Foreground Object Transformation}                      

\author{Chaofei Wang}[orcid=0000-0002-3678-691X]
\author{Shiji Song}
\cormark[1]
\author{Qisen Yang}
\author{Xiang Li}
\author{Gao Huang}

\address{Department of Automation, Tsinghua University, Beijing, China}

\cortext[cor1]{indicates corresponding author}

\begin{abstract}
Traditional fine-grained image classification generally requires abundant labeled samples to deal with the low inter-class variance but high intra-class variance problem. However, in many scenarios we may have limited samples for some novel sub-categories, leading to the fine-grained few shot learning (FG-FSL) setting. To address this challenging task, we propose a novel method named foreground object transformation (FOT), which is composed of a foreground object extractor and a posture transformation generator. The former aims to remove image background, which tends to increase the difficulty of fine-grained image classification as it amplifies the intra-class variance while reduces inter-class variance. The latter transforms the posture of the foreground object to generate additional samples for the novel sub-category. As a data augmentation method, FOT can be conveniently applied to any existing few shot learning algorithm and greatly improve its performance on FG-FSL tasks. In particular, in combination with FOT, simple fine-tuning baseline methods can be competitive with the state-of-the-art methods both in inductive setting and transductive setting. Moreover, FOT can further boost the performances of latest excellent methods and bring them up to the new state-of-the-art. In addition, we also show the effectiveness of FOT on general FSL tasks.
\end{abstract}

\begin{keywords}
fine-grained \sep few shot learning \sep foreground object transformation \sep image classification \sep saliency map matching
\end{keywords}

\maketitle

\section{Introduction}
\label{sec:intro}

\begin{figure}[t]
	\vspace{2ex}
	\begin{center}
		\includegraphics[width=0.9\linewidth]{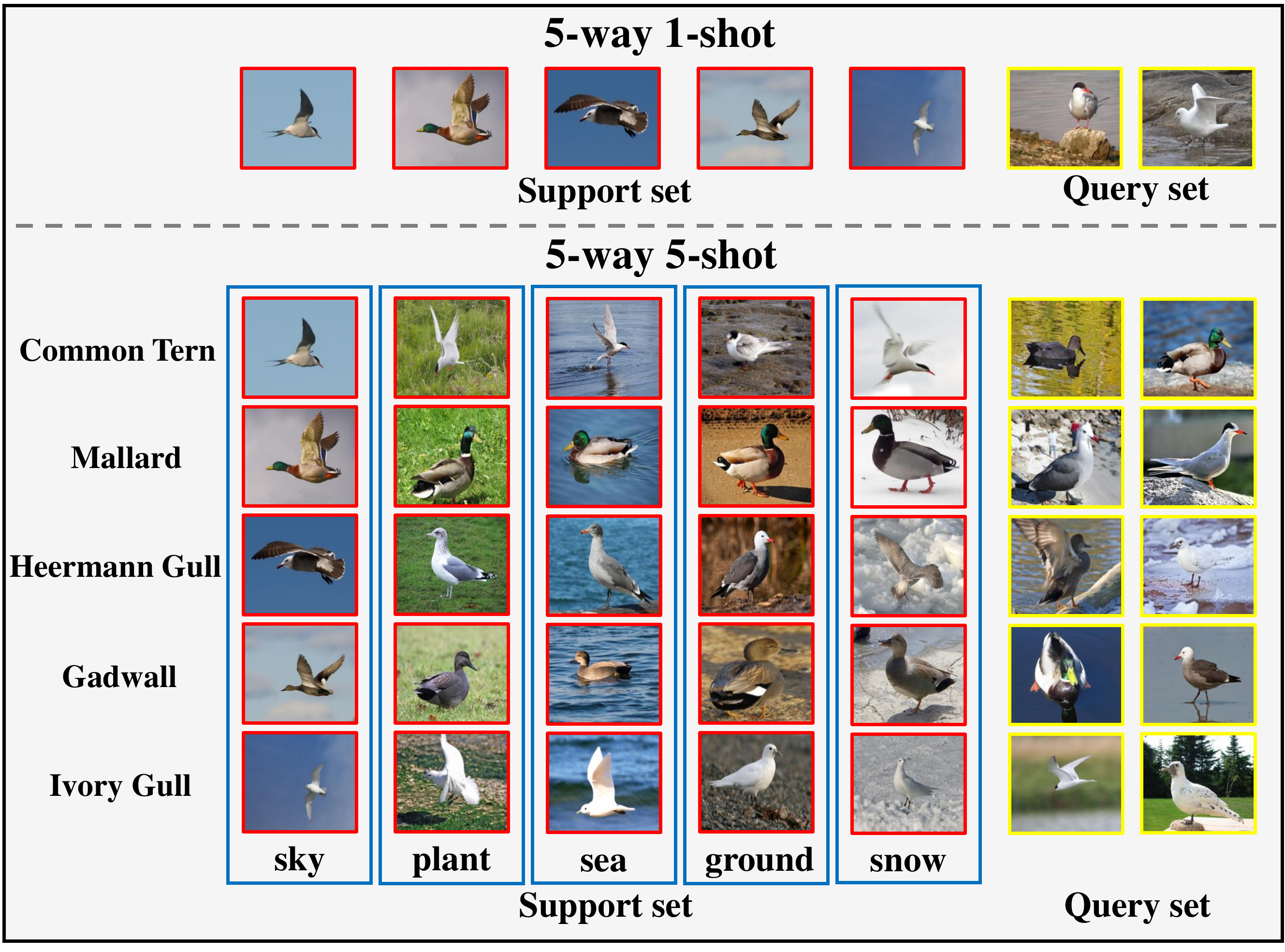}
	\end{center}
	\caption{Illustration of the FG-FSL task. The upper part is the 5-way 1-shot task, and the lower part is the 5-way 5-shot task. In the lower part, rows represent different bird species: Common Tern, Mallard, Heermann Gull, Gadwall, Ivory Gull, and columns represent birds with different backgrounds: sky, plant, sea, ground, snow. Observing horizontally, every category has high intra-class variance. Observing longitudinally, different categories have low inter-class variance. That is the novel difficulty from FSL to FG-FSL.}
	\vspace{-2ex}
	\label{fig:FGFSL}
\end{figure}

As a popular and challenging problem in computer vision, fine-grained image classification has been an active research area for several decades \cite{wei2019deep}. The goal is to recognize images belonging to multiple sub-categories of a super-category \cite{zhong2018multi} e.g., different species of animals, different models of cars, different kinds of retail products, etc. With the fast development of deep learning, fine-grained image classification has made a significant leap forward, typically relying on supervised learning from large amounts of labeled samples \cite{wei2018mask,huang2016part,xiao2015application,zhao2017diversified,hu2019see,fu2017look}. In many real-world scenarios, however, it may happen that very sparse training samples are available for some sub-categories. For example, biologists often discover rare bird or fish as new species, and car makers always produce new models of cars. This leads to a more challenging setting, namely the fine-grained few shot learning (FG-FSL) problem as shown in \autoref{fig:FGFSL}. 

To effectively learn from few samples, many few shot learning (FSL) algorithms have been proposed in recent years \cite{finn2017model,snell2017prototypical,vinyals2016matching,sung2018learning,dhillon2019baseline,boudiaf2020transductive}. However, few of them focus on the FG-FSL task. Besides the characteristics of general FSL tasks, FG-FSL also inherits the difficulty of fine-grained image classification tasks, which appears as low inter-class variance but high intra-class variance. There are two main challenges in the FG-FSL task. First, subtle features for distinguishing different sub-categories always reside in the foreground object, but there exist insufficient samples to learn such discriminative features for novel sub-categories. Second, the backgrounds of a same sub-category are quite different (e.g. each row in the lower part of \autoref{fig:FGFSL}) but backgrounds of different sub-categories may appear similar (e.g. each column in the lower part of \autoref{fig:FGFSL}). It means that the image background tends to play a negative role in the FG-FSL task, as it drastically increases the intra-class variance and reduces the inter-class variance.
Many data augmentation methods have been proposed to alleviate the first problem \cite{hariharan2017low,schwartz2018delta,zhang2018metagan,chen2019multi,chen2019image,gao2018low}. However, most of them take the image as a whole in the pixel space or feature space, which could not avoid the negative influence of background while generating additional training samples. Few attention based methods have been proposed to solve the second problem \cite{jiang2020few,hou2019cross} in FSL tasks, which rely on complex network structures and training strategies, because it is very challenging to localize foreground objects with only image-level labels and very few samples. In fact, it is even more difficult to solve the two problems simultaneously.

\begin{figure}[t]
	\begin{center}
		\includegraphics[width=0.95\linewidth]{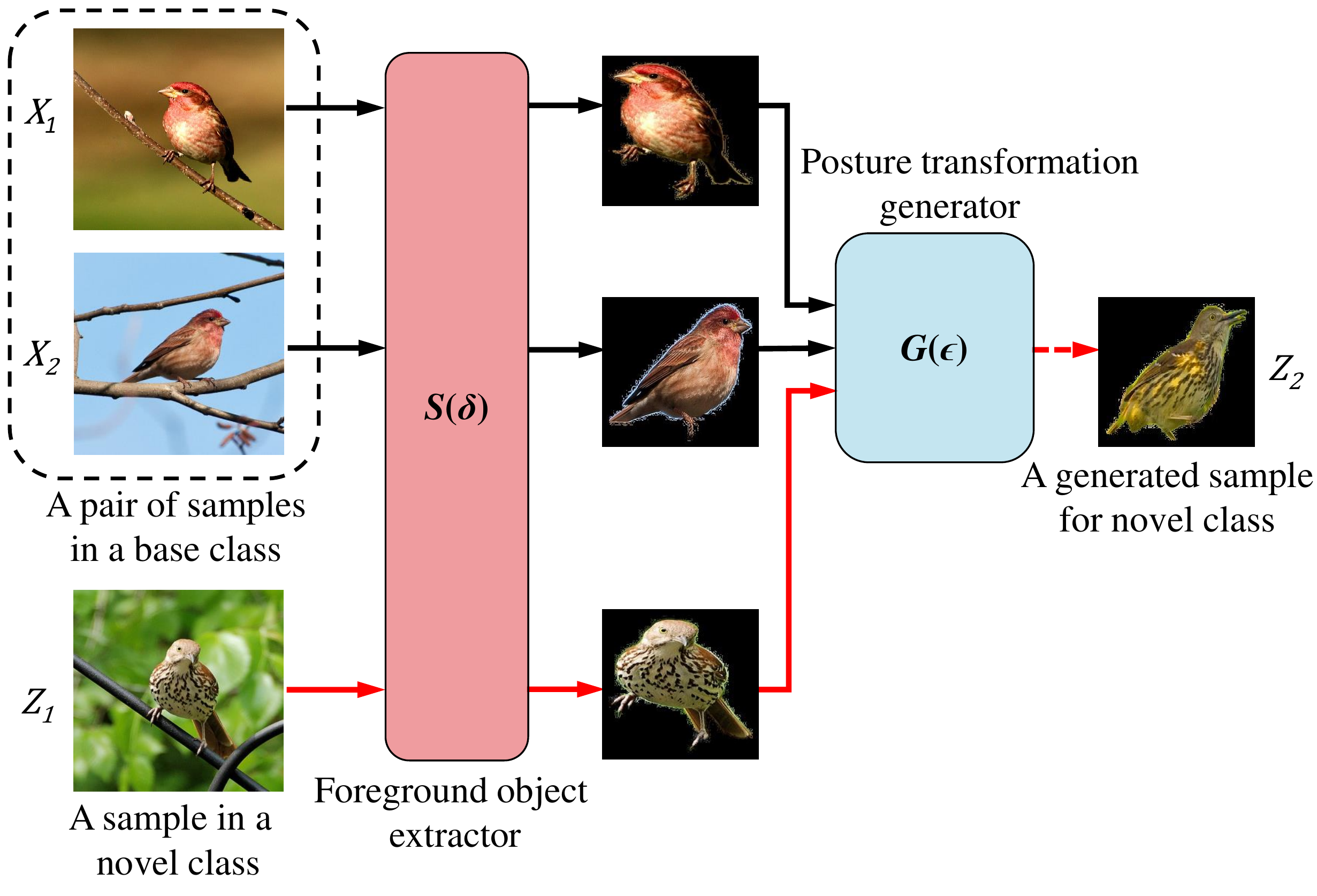}
	\end{center}
    \vspace{-2ex}
	\caption{A brief overview of proposed FOT method. For a sample $Z_{1}$ in a novel class, we find a pair of samples $(X_{1}, X_{2})$ in a base class, and use a foreground object extractor to remove backgrounds of these samples. Then, the posture transformation between $X_{1}$ and $X_{2}$ is added to $Z_{1}$ by a generator, in order to get an additional sample like $Z_{2}$ for the novel class.}
	\vspace{-2ex}
	\label{fig:keyidea}
\end{figure}

In this paper, we propose to consider the foreground and background separately, and introduce a foreground object transformation (FOT) method for FG-FSL tasks. It mainly includes a foreground object extractor and a posture transformation generator, which correspond to background remove and foreground augmentation respectively. Specifically, the first component leverages the popular salient object detection (SOD) method to identify the foreground and background area in an image, and then extract a zoomed-in version of the foreground object without background information. This avoids the negative effect of the background and reinforces the features of the foreground object. The second component contains a generator to learn the posture transformations of foreground objects in base sub-categories (hereafter referred to as base classes, which generally have many training samples), and then transform the posture of foreground objects to generate additional samples with different postures for novel sub-categories (hereafter referred to as novel classes, which generally have very few training samples). A brief overview of FOT is shown in \autoref{fig:keyidea}. Specially, we propose a saliency map matching strategy to construct a quadruplet-based training set for the posture transformation generator learning.  

Based on a common fine-tuning baseline, Baseline++ \cite{chen2019closer}, FOT can be extended as a complete method to handle FG-FSL tasks independently. On fine-grained benchmark datasets, extensive experiments show that FOT significantly outperforms existing inductive inference methods, including typical FSL \cite{finn2017model,snell2017prototypical,vinyals2016matching,sung2018learning}, FG-FSL \cite{wei2019piecewise,huang2019compare,li2019distribution,li2019revisiting}, and hallucination based FSL methods \cite{hariharan2017low, schwartz2018delta, wang2018low}. Based on a transductive fine-tuning baseline \cite{dhillon2019baseline}, FOT can also be extended as an independent transductive inference method, which is competitive with the state-of-the-art transductive inference methods \cite{qiao2019transductive,ziko2020laplacian,wang2020instance,boudiaf2020transductive}. Furthermore, as a data augmentation method, FOT is conveniently applied to any existing FSL method and improves its performance on FG-FSL tasks. In combination with FOT, several latest methods are boosted to the new state of the art. In addition, we also verify the generalization capability of FOT on general FSL tasks, and conduct the ablation experiment to analyse the effectiveness of different components.

The main contributions of our work are:
\begin{itemize}[itemsep= 0 pt,topsep = 0 pt, parsep = 0 pt]
	\item We propose a novel data augmentation method, foreground object transformation (FOT), which enhances the diversity of foreground while eliminating the negative effect of background by considering the foreground and background separately.
	\item We propose a novel saliency map matching strategy to construct a quadruplet-based dataset to train a posture transformation generator, which can effectively transform the posture of foreground objects in novel classes, yielding diversified and visualized augmented images without changing their class labels.
	\item Our method can be conveniently applied to any existing FSL algorithm to handle FG-FSL tasks effectively. In combination with FOT, simple fine-tuning baseline methods can obtain competitive performance with the state-of-the-art methods, and the state-of-the-art methods can be pushed to new heights.
\end{itemize}

\section{Related work}

We briefly review existing research on related topics.

\subsection{Few shot learning}

The human visual systems can recognize novel classes with extremely few labeled samples. It is thus of great interest for neural networks to learn to recognize novel classes with a few labeled samples, known as few shot learning (FSL). Currently meta-learning has been a broad paradigm for FSL tasks. Most of popular works can be divided into three main categories: initialization based, metric learning based, and hallucination based methods. Initialization based methods aim to learn good model initialization (i.e., the parameters of a network) so that the classifier for novel classes can be learned with a few labeled samples and a few gradient updated steps \cite{nichol2018reptile,finn2017model,finn2018probabilistic,rusu2018meta}. Metric learning based methods aim to learn a sophisticated comparison model to determine the similarity of two images \cite{lim2021efficient,vinyals2016matching,snell2017prototypical,sung2018learning,liu2020meta,garcia2017few}. Hallucination based methods learn a generator from samples in the base classes and use the learned generator to hallucinate new novel class samples for data augmentation\cite{hariharan2017low,schwartz2018delta,wang2018low,chen2019multi,chen2019image}. According to this classification criterion, the proposed FOT belongs to the hallucination based methods. 

The most relevant to our approach is the work of \cite{hariharan2017low}, which considers the image as a whole and conjectures that the relative linear offset in feature space between a pair of samples in the same class conveys information on a valid deformation. However, \cite{hariharan2017low} is difficult to apply to small fine-grained datasets. The main reason is that, even within the same class, the deformations of images are quite complex, including color, posture, background, size of object, and so on. In order to learn these infinite deformations accurately, it theoretically requires infinite data and sufficiently complex generative models. Furthermore, although \cite{hariharan2017low} improves the FSL performance to some extent, the generated samples could not be visualized and accurately indicate the learned deformation style. Different from \cite{hariharan2017low}, we remove the negative effect of the background by using a foreground object extractor, and then employ saliency maps to constrain image deformations to the posture transformation in pixel space, which greatly reduces the cost of the generator training and results in visual generated samples.

Recently, some approaches tackle FSL problems by resorting to additional unlabeled data \cite{ren2018meta,li2019learning,wang2020instance,liu2018learning,liu2019prototype,boudiaf2020transductive}. Specifically, semi-supervised FSL methods \cite{ren2018meta,li2019learning,wang2020instance} enable unlabeled data from the same categories to better handle the true distribution of each class. Furthermore, transductive inference methods \cite{liu2018learning,qiao2019transductive,ziko2020laplacian,wang2020instance,boudiaf2020transductive}, which utilize the unlabeled samples from the query set, show great performance improvements over inductive inference. As a data augmentation method, the proposed FOT is adapted to both inductive inference methods and transductive inference methods. In combination with FOT, some excellent transductive inference methods \cite{wang2020instance,boudiaf2020transductive} can be brought up to the new state of the art on FG-FSL tasks.

\subsection{Fine-grained image classification}

Fine-grained image classification is a challenging problem and has been an active topic \cite{wei2019deep}. Since subtle visual differences mostly reside in local regions of images, discriminative part localization is crucial for fine-grained image classification. There are numerous emerging works proceeding along part localization \cite{wei2018mask,zhang2014part,zhang2016spda,huang2016part}, which tend to learn accurate part localization models with manual object bounding boxes and part annotations. Considering that the annotations are laborious and expensive, some researchers begin to focus on how to exploit parts under a weakly-supervised setting with only image-level labels \cite{ge2019weakly,zhang2019part,zhang2015weakly,he2018fast,zhang2016weakly,xiao2015application}. Additionally, some weakly-supervised methods use visual attention mechanism to automatically capture the informative regions \cite{hu2018weakly,ji2019attention,sermanet2014attention,xiao2015application,zhao2017diversified,hu2019see,fu2017look,zheng2017learning,wei2017selective,choe2020attention}. Compared with previous work, we study fine-grained image classification in a challenging few shot learning setting. We build the classifier of novel classes using few samples with only image-level labels, which belongs to the weakly-supervised methods typically. \cite{wei2019piecewise} proposed the first FG-FSL model, which adopted a piecewise mappings function in the classifier mapping module to improve generalization. \cite{li2020bsnet} proposed to employ two similarity measures in the metric learning based methods, generating more discriminative features than using a single measure.

\subsection{Salient object detection}

A salient object detector highlights the image region containing foreground objects which correlate with human visual attention, thus producing a dense likelihood saliency map which assigns some relevance score in range $[0, 1]$ to each pixel. With the success of deep learning in computer vision, more and more deep learning based SOD methods have been springing up since 2015 \cite{li2015visual,wang2015deep,zhao2015saliency}. Earlier deep SOD models typically utilize multi-layer perceptron classifiers to predict the saliency score of deep features extracted from each image processing unit \cite{li2015visual,wang2015deep,zhao2015saliency}. Inspired by the great success of Fully Convolutional Network (FCN) \cite{long2015fully} in semantic segmentation, latest deep SOD methods adapt popular classification models, e.g., VGGNet \cite{simonyan2014very} and ResNet \cite{he2016deep}, into fully convolutional ones to directly output saliency maps instead of classification scores. These deep SOD methods benefit from end-to-end spatial saliency representation learning and efficiently predict saliency maps in a single feed-forward process \cite{chen2018reverse,li2018contour,liu2018picanet,wang2018detect,qin2019basnet}. 

A pre-trained SOD model can identify the foreground and background of an image automatically, which gives us an inspiration to deal with FG-FSL tasks. We need to choose a pre-trained SOD model with good generalization capability, which means that the SOD model performs well on fine-grained benchmark datasets, even though it is pre-trained on a disjoint SOD dataset. We compare the performance of 
three code-exposed methods, including PiCANet \cite{liu2018picanet}, DGRL \cite{wang2018detect} and BASNet \cite{qin2019basnet}. BASNet \cite{qin2019basnet} is chosen as our pre-trained SOD model to capture the foreground object of images, because it has excellent generalization capacity and gets clearer, sharper saliency maps.

\section{Method}

In this section, we first review the fine-tuning baselines in the inductive and transductive setting for FSL tasks in Section \ref{sec:baseline}. Then, we present the architecture of FOT and elaborate its algorithm in Section \ref{sec:FOT}. Next, we explain the two novel components of FOT, foreground object extractor and posture transformation generator, in section \ref{sec:foerground_extractor} and section \ref{sec:posture_transform} respectively.

\subsection{Review the fine-tuning baselines}
\label{sec:baseline}
\begin{figure}[h]
	\begin{center}
		\includegraphics[width=0.9\linewidth]{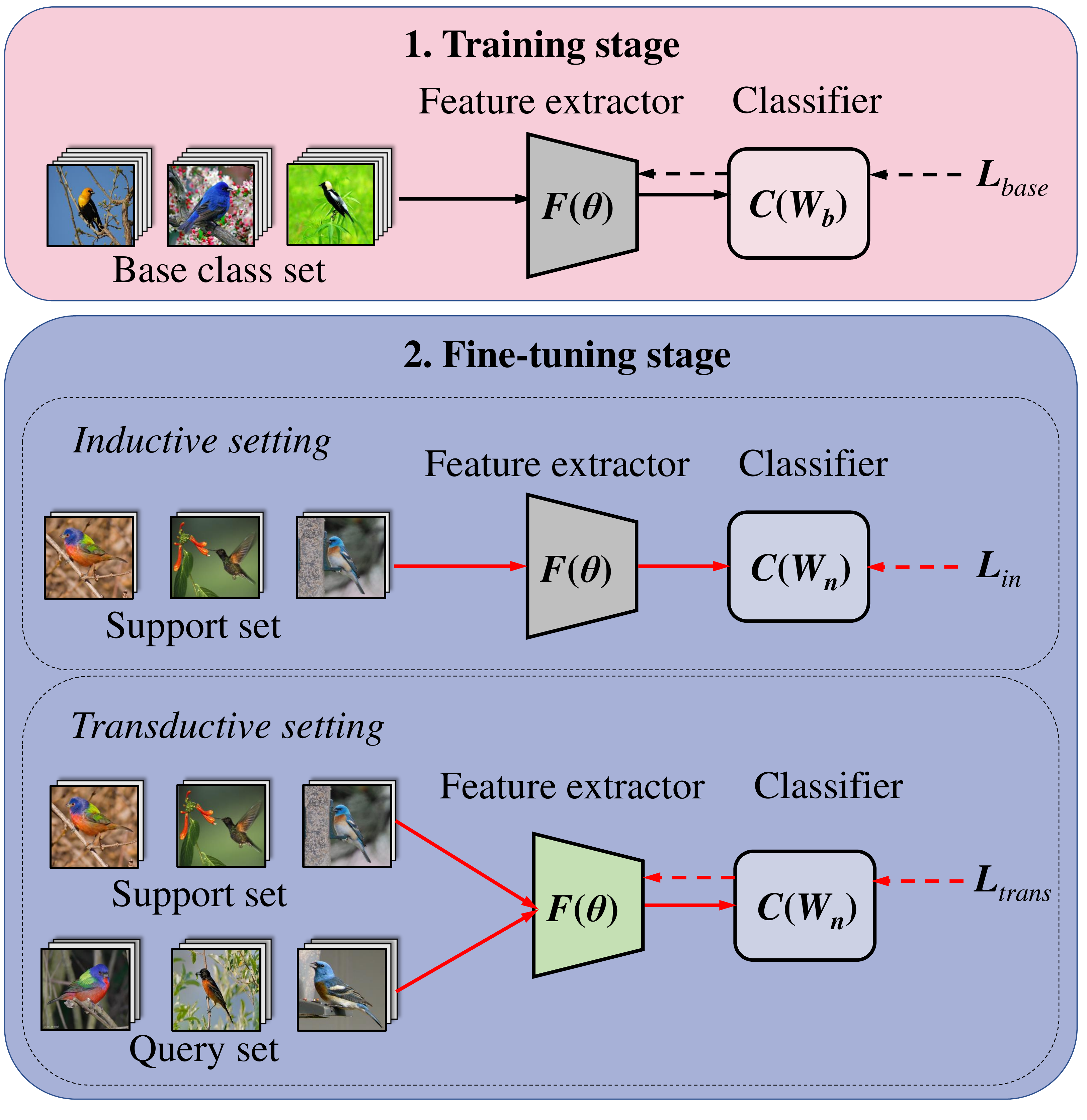}
	\end{center}
	\caption{Fine-tuning baseline methods for FG-FSL tasks. Both the inductive inference baseline and the transductive inference baseline have two stages: training stage and fine-tuning stage. Differences between them are in the fine-tuning stage, including different input data, different loss functions and different gradient propagation paths. The solid lines represent the data transmission paths and the dashed lines represent the gradient propagation paths.}
	\vspace{-2ex}
	\label{fig:baseline}
\end{figure}
Given a base class set $D_{b}$ with abundant labeled samples, a novel class support set $D_{s}$ with few labeled samples, and a novel class query set $D_{q}$ with limited unlabeled samples, the goal of FSL algorithms is to train classifiers for novel classes and test its classification accuracy on the query set. A FSL baseline method generally follows the standard transfer learning procedure of network pre-training and fine-tuning \cite{chen2019closer,dhillon2019baseline}, which can be directly transferred to FG-FSL tasks as shown in \autoref{fig:baseline}. In the training stage, a feature extractor $F(\theta)$ and a classifier of base classes $C(W_{b})$ are trained with samples in $D_{b}$ by minimizing a standard cross-entropy loss $L_{base}$. It is formalized as follows:
\begin{equation}
	L_{base} = -\frac{1}{N_{b}}\sum_{(x,y)\in D_{b}}y\cdot  log(p(x)),
	\label{equ:baseloss}
\end{equation}
where (x,y) represents a sample x with true label y, $N_{b}$ is the size of mini-batch from $D_{b}$, $p(x)$ is the softmax output of the classifier $C(W_{b})$. In the fine-tuning stage, we consider two different schemes: inductive inference\cite{chen2019closer} and transductive inference\cite{dhillon2019baseline}. In inductive setting, we fix the feature extractor $F(\theta)$ and train a new classifier $C(W_{n})$ with samples in $D_{s}$ by minimizing a standard cross-entropy loss $L_{in}$, which is formalized as follows: 
\begin{equation}
	L_{in} = -\frac{1}{N_{s}}\sum_{(x,y)\in D_{s}}y\cdot  log(p(x)),
	\label{equ:inloss}
\end{equation}
where $N_{s}$ represents the number of samples in the support set. In transductive setting, we fine-tune the feature extractor $F(\theta)$ and train the new classifier $C(W_{n})$ with samples in $D_{s}$ and $D_{q}$ by minimizing an expanded loss $L_{trans}$, which is formalized as follows:
\begin{small}
\begin{equation}
	L_{trans} = -\frac{1}{N_{s}}\!\sum_{(x,y)\in D_{s}}\!y\cdot  log(p(x))+\frac{1}{N_{q}}\!\sum_{x\in D_{q}}\!p(x)\cdot  log(p(x)),
	\label{equ:transloss}
\end{equation}
\end{small}
where $N_{q}$ represents the number of samples in the query set.

We clarify that these two baseline methods are not our contribution. The inductive inference baseline method has been extensively studied in \cite{chen2019closer} and the transductive inference baseline method has been proposed in \cite{dhillon2019baseline}. Typically, the feature extractor $F(\theta)$ is a ConvNet-4 or ResNet-18 backbone, and the classifiers $C(W_{b})$ and $C(W_{n})$ are cosine-distance classifiers by following \cite{chen2019closer,dhillon2019baseline}.

\subsection{Foreground object transformation}
\label{sec:FOT}

To solve the two key problems of FG-FSL tasks mentioned in Section \ref{sec:intro}, we consider the foreground and background separately and propose two novel components, a foreground object extractor and a posture transformation generator. The former aims to extract a zoomed-in version of the foreground object from an input sample. It avoids the negative effect of the background and highlights the features of the foreground object. The latter is used to generate additional samples for each novel class by transforming the posture of extracted foreground object. We add these two components to a fine-tuning baseline and form a new FG-FSL method, which is called foreground object transformation (FOT). In fact, combining with different baselines in Section \ref{sec:baseline}, we can get an inductive FOT or a transductive FOT. For simplicity, we mark the transductive FOT as FOT$^{*}$ in this paper. On the other hand, since the application of our method is independent of inductive or transductive setting, we do not specifically discuss FOT$^{*}$ except for the experimental part. 

\begin{figure*}
	\begin{center}
		\includegraphics[width=0.9\linewidth]{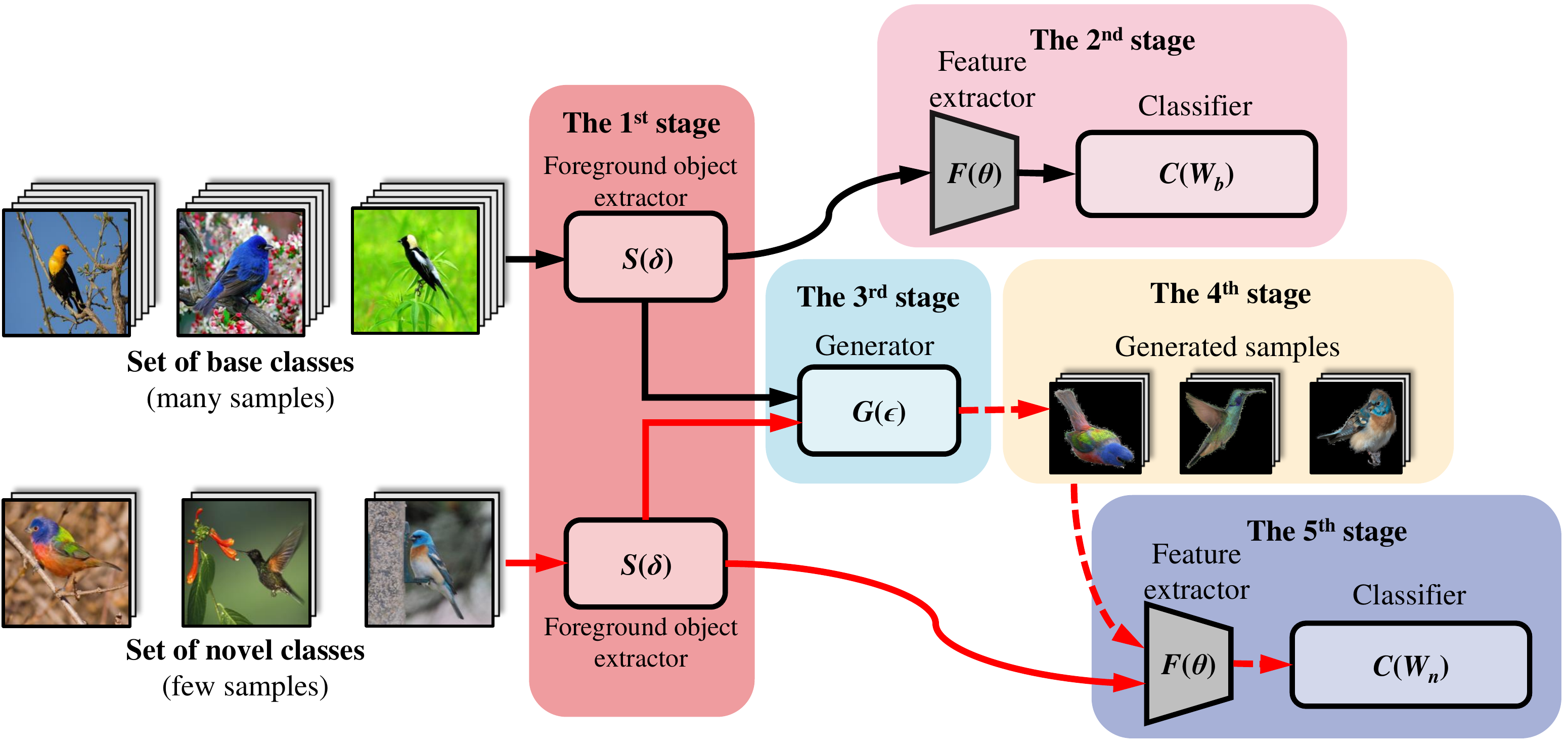}
	\end{center}
	\caption{The architecture of the proposed FOT method. Its algorithm is split into five steps shown in different color boxes. Based on a fine tune baseline method, we add two additional components: a foreground object extractor $S(\delta)$ and a posture transformation generator $G(\epsilon)$.}
	\vspace{-2ex}
	\label{fig:SHWB}
\end{figure*}

The architecture of FOT is shown in \autoref{fig:SHWB}. To illustrate how FOT works in detail, we split the algorithm into five steps. \autoref{fig:SHWB} shows the procedure and relevant notations in different color boxes.

\textbf{Extracting foreground object (1$^{st}$ stage).} We use a foreground object extractor $S(\delta)$ to obtain the foreground object of an input image. More details on the foreground object extractor are provided in Section \ref{sec:foerground_extractor}. All samples from both base classes and novel classes are processed in this way. Subsequent steps use the processed samples instead of original images.

\textbf{Training on base classes (2$^{nd}$ stage).} We train the feature extractor $F(\theta)$ and the classifier of base classes $C(W_{b})$ with Equ. \ref{equ:baseloss} by feeding with samples of base classes.

\textbf{Learning posture transformation (3$^{rd}$ stage).} 
In order to transform the posture of foreground object, we design a generator  $G(\epsilon)$ to learn the posture transformations of foreground objects from base classes. We adopt a saliency map matching strategy to construct a quadruplet-based dataset for training the generator $G(\epsilon)$. More details on how to construct the dataset and design the generator structure are provided in Section \ref{sec:posture_transform}. 

\textbf{Generating samples for novel classes (4$^{th}$ stage).} 
Using the trained posture transformation generator $G(\epsilon)$, we can transform the posture of foreground objects from the novel classes to obtain more samples with different postures. In this way, the support set can be effectively augmented.

\textbf{Training on novel classes (5$^{th}$ stage).} 
We fix the feature extractor $F(\theta)$ and train the classifier $C(W_{n})$ for novel classes with samples in the augmented support set by minimizing Equ. \ref{equ:inloss}. Please note that if it is FOT$^{*}$, we fine-tune the feature extractor $F(\theta)$ and train the new classifier $C(W_{n})$ with samples in the augmented support set and query set by minimizing Equ. \ref{equ:transloss}.

Through the above steps, we get the specific parameters of all components. In the test stage, we pass test samples through the foreground object extractor $S(\delta)$, feature extractor $F(\theta)$ and novel class classifier $C(W_{n})$ to get the final classification labels. 

\subsection{Foreground object extractor}
\label{sec:foerground_extractor}

As previously mentioned, image background plays a negative role in FG-FSL tasks, because it tends to amplify the intra-class variance while reduce inter-class variance. Therefore, we propose to construct a foreground object extractor to remove the background and capture the foreground object of an image. 

Fortunately, the rapid development of SOD technology has made it possible to identify the background and foreground of an image. We propose to choose a pre-trained SOD model to construct the foreground object extractor. We select three code-exposed supervised SOD models, PiCANet \cite{liu2018picanet}, DGRL \cite{wang2018detect} and BASNet \cite{qin2019basnet}, trained on DUTS \cite{wang2017learning} (the largest SOD dataset containing 10,553 training and 5,019 test images, which is disjoint with fine-grained datasets). Testing the performance of these methods on fine-grained datasets, we choose BASNet \cite{qin2019basnet} as our SOD model due to its superior performance, which means the saliency maps produced by BASNet are clearer and sharper than others on these datasets.
\begin{figure*}
	\centering
	\includegraphics[width=0.9\linewidth]{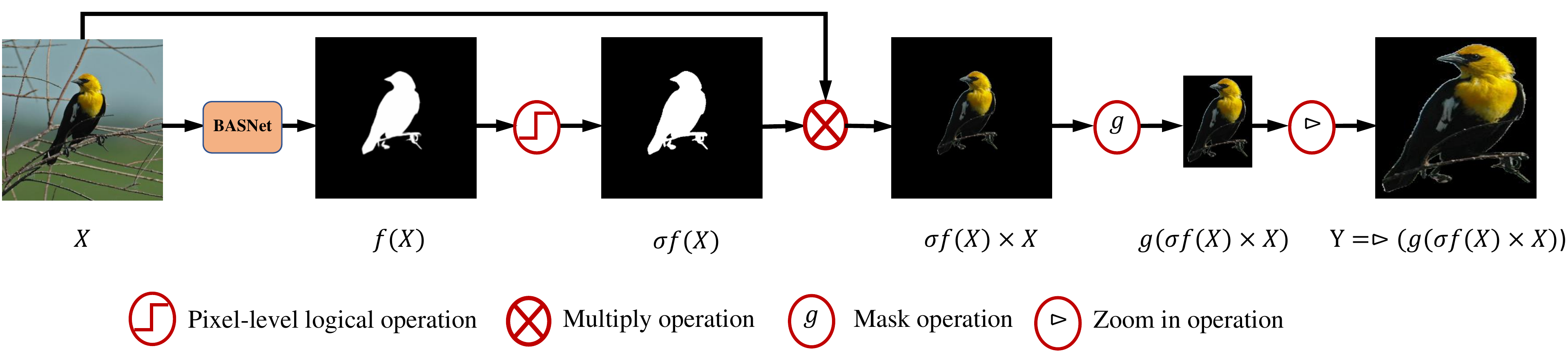}
	\caption{The detailed architecture of the foreground object extractor. The evolution from $X$ to $Y$ is introduced step by step.}
	\vspace{-2ex}
	\label{fig:s1}
\end{figure*}

The architecture of foreground object extractor $S$ is constructed with a pre-trained BASNet module $f$, a pixel-level logical operation $\sigma$, a multiply operation $\otimes$, a mask operation $g$ and a zoom-in operation $\vartriangleright$ as shown in \autoref{fig:s1}. In the experiment, given an input picture $X$, let $Y$ be the output of $S$. The process from $X$ to $Y$ is formally described as follows:
\begin{equation}
	\begin{aligned}
		\quad
		\begin{cases}
			X^{s}=f(X), \\
			\widehat{X^{s}}=\sigma(X^{s}), \\
			X^{m}=g(\widehat{X^{s}}\otimes X), \\
			Y=\vartriangleright(X^{m}). \\
		\end{cases}
	\end{aligned}
	\label{equ:equation1}
\end{equation}
Specifically, given an image $X$ with shape $C\times H\times W$, we first obtain the original saliency map $X^{s}$ via the BASNet module $f$. Considering that $X^{s}$ is still an rgb-image with pixel value between $[0,255]$, we simply apply a pixel-level logical operation $\sigma$ as follows:
\begin{equation}
	\begin{aligned}
		\widehat{X^{s}}=\sigma(X^{s})\Leftrightarrow\widehat{X^{s}_{i}}=
		\begin{cases}
			1,\ \overline {X^{s}_{i}} \geqslant \beta, \\
			0,\  \overline {X^{s}_{i}}<\beta, \\
		\end{cases}
	\end{aligned}
\end{equation}
where $\widehat{X^{s}_{i}}$ is the value of the $i^{th}$ pixel in $\widehat{X^{s}}$ which is a one-channel logical map with shape $1\times H\times W$, $\overline {X^{s}_{i}}$ is the mean of $X^{s}_{i}$ on channel $C$, and $\beta$ is a threshold value. Simply, we set $\beta$ as $40$ for all datasets. Then, we do multiplication between $X$ and $\widehat{X^{s}}$ to get a separate foreground object with black background. Furthermore, we crop $\widehat{X^{s}}\otimes X$ to get the part of foreground object $X^{m}$ with a mask operation $g$, which captures the position and size of a bounding box according to the values of $\widehat{X^{s}_{i}}$. Finally, we zoom in the part $X^{m}$ with a zoom-in operation $\vartriangleright$, and then get the zoomed-in version $Y$ of the foreground object. Another brief equation equivalent to \autoref{equ:equation1} is as follows:
\begin{equation}
	\begin{aligned}
		Y=\vartriangleright(g(\sigma(f(X))\otimes X)).
	\end{aligned}
	\label{equ:equation2}
\end{equation}

\subsection{Posture transformation generator}
\label{sec:posture_transform}
It is difficult to estimate the true distribution of a novel class with high intra-class variance by utilizing limited samples. For example, if the novel class is a particular bird species, then we may only have a few samples of the bird perched on a branch, but none in flight. The classifier might erroneously conclude that this novel class only consists of perched birds.

However, this mode of posture transformation is common to many other bird species in the base classes. From the base class samples, we can learn the posture transformation from perched birds to flying birds. Then we may apply this transformation to a perched bird in a novel class to generate a flying bird for the novel class. Similarly, many different posture transformations can be learned from base classes and applied to novel classes. In this way, the diversity of novel class samples is significantly increased, which is beneficial to the generalization of the classifier.

\begin{figure}[t]
	\begin{center}
		\includegraphics[width=0.9\linewidth]{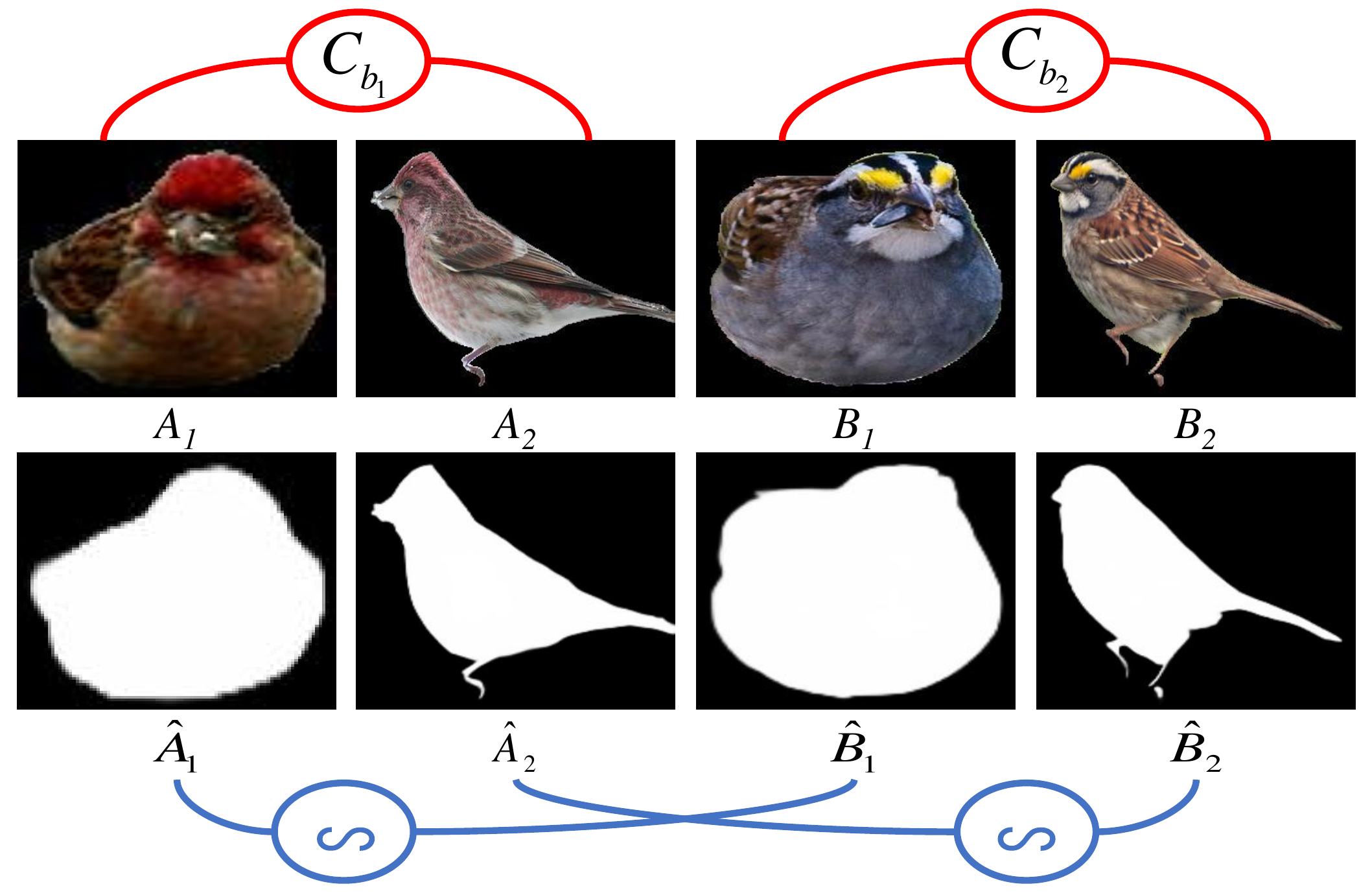}
	\end{center}
	\caption{The internal relationship of a quadruplet $\{A_{1}, A_{2},$ $B_{1}, B_{2}\}$ in $D_{g}$.}
	\vspace{-2ex}
	\label{fig:s3}
\end{figure}

\begin{figure*}
	\begin{center}
		\includegraphics[width=0.9\linewidth]{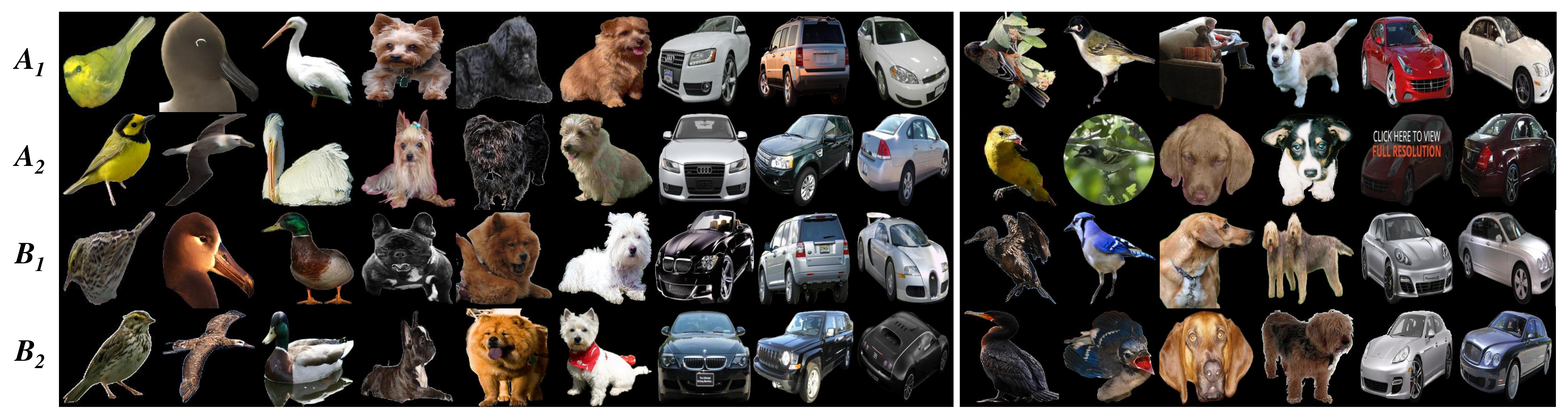}
	\end{center}
	\caption{Some examples of quadruplets from $D_{g}$. Each column represents a quadruplet $\{A_{1}, A_{2}, B_{1}, B_{2}\}$. The left side shows some positive examples while the right side shows some negative ones.}
	\vspace{-2ex}
	\label{fig:s4}
\end{figure*}

\textbf{Construction of training set.} To learn posture transformations in base classes, we first construct an additional dataset $D_{g}$ with a large number of quadruplets like $\{A_{1}$, $A_{2}$, $B_{1}$, $B_{2}\}$. The internal relationship of a quadruplet $\{A_{1}$, $A_{2}$, $B_{1}$, $B_{2}\}$ has been shown in \autoref{fig:s3}. Specifically, $(A_{1}, A_{2})$ is a pair of samples in one base class, $(B_{1}, B_{2})$ is another pair of samples in another base class. $\hat{A_{1}},\hat{A_{2}},\hat{B_{1}},\hat{B_{2}}$ represent respectively the saliency maps of $A_{1}, A_{2}, B_{1}, B_{2}$. We make sure that $\hat{A_{1}}$ is similar to $\hat{B_{1}}$ while $\hat{A_{2}}$ is similar to $\hat{B_{2}}$. Since the saliency map represents the posture of foreground object, the posture transformation of $A_{1}$ to $A_{2}$ is similar to that of $B_{1}$ to $B_{2}$. Some real examples of quadruplets from $D_{g}$ have been shown in \autoref{fig:s4}. Obviously, it verifies an objective rule that similar postures are often accompanied by similar saliency maps. However, the opposite is not always true. The right side shows some counter-examples. In these cases, the saliency maps may be similar in different postures, especially in dog and car datasets. In fact, the negative examples are relatively rare in $D_{g}$, so we simply ignore them. We can obtain many quadruplets with these constrains by searching within base classes. The seaching strategy is called \textbf{saliency map matching}, which is formalized as follows: 
\begin{equation}
	\begin{aligned}
		& find\ all \ \{A_{1}, A_{2}, B_{1}, B_{2}\}, \\
		&s.t.\quad
		\begin{cases}
			A_{1}, A_{2} \in C_{b_{1}}, \\
			B_{1}, B_{2} \in C_{b_{2}}, \\
			Distance(\hat{A_{1}},\hat{B_{1}})<\alpha, \\
			Distance(\hat{A_{2}},\hat{B_{2}})<\beta, \\
			b_{1}\neq b_{2},\ \alpha>0,\ \beta>0.\\
		\end{cases}
	\end{aligned}
\end{equation}
Where, $C_{b_{1}}$, $C_{b_{2}}$ represent two different base classes, $\hat{A_{1}}$, $\hat{A_{2}}$, $\hat{B_{1}}$, $\hat{B_{2}}$ represent respectively the saliency maps of $A_{1}$, $A_{2}$, $B_{1}$, $B_{2}$. Euclidean distance is used to calculate the distance between two saliency maps. It is non-trivial to determine the values of $\alpha,\beta$. Therefore, we simply choose the top 5 closest $ B_{1}$ for each $A_{1}$, and then choose the closest $ B_{2}$ for each $A_{2}$. Sufficient quadruplets can be acquired to avoid overfitting by traversing all base classes.

when training the posture transformation generator $G(\epsilon)$, we concatenate $\{A_{1}, A_{2}, B_{1}\}$ as input, take $B_{2}$ and $\widetilde{B_{2}} = G_{\epsilon}([A_{1}, A_{2}, B_{1}])$ as the target and predicted output.  


\textbf{Design of generator.} We simply use an encoder network consisting of three convolutional layers with built-in resblocks and symmetric deconvolutional layers for the generator. In order to ensure that the generated samples play a positive role in the classification, we connect the trained feature extractor $F(\theta)$ and classifier $C(W_{b})$ to the output layer of the encoder. For each quadruplet $\{A_{1}, A_{2}, B_{1}, B_{2}\}$, we minimize the following loss function:
\begin{equation}
	L_{g} = \lambda L_{mse}(\widetilde{B_{2}}, B_{2}) + L_{ce} (W_{b},F_{\theta}(\widetilde{B_{2}}), y),
\end{equation}
where $L_{mse}(\widetilde{B_{2}}, B_{2})$ is the mean squared error between $\widetilde{B_{2}}$ and $B_{2}$. 
$L_{ce} (W_{b},F_{\theta}(\widetilde{B_{2}}), y)$ is the cross-entropy classification loss of the classifier $C(W_{b})$ on the sample $(\widetilde{B_{2}},y)$, where $W_{b}$ is the fixed classifier of base classes trained before, $F(\theta)$ is the fixed feature extractor trained before, $y$ is the label of $B_{2}$, $\lambda$ is a regulated parameter.

\textbf{Sample generation for novel classes.} For each sample $Z_{1}$ from a novel class, we find some samples $X_{1}$ in base classes with similar posture, which means that euclidean distance between the saliency map of $X_{1}$ and that of $Z_{1}$ is minimum. Then we randomly sample some pairs $(X_{1},X_{2})$ which represent posture transformations of $X_{1}$ to $X_{2}$ from $D_{g}$. Fed with $\{X_{1}, X_{2}, Z_{1}\}$, the posture transformation generator is able to generate $\widetilde{Z_{2}}$ with the posture of $X_{2}$ and the class feature of $Z_{1}$. Theoretically, we can get many $X_{1}$ with similar posture from base classes, so that many generated $\widetilde{Z_{2}}$ can be obtained. A right amount of generated samples are beneficial to increase the diversity of a novel class. Too many generated samples will lead to additional bias of the true distribution. Therefore, we have a hyperparameter $k$, which is an appropriate number of generated samples for each novel class. Empirically, in order to ensure that the generated samples are not dominant, $k$ usually does not exceed 3 for 1-shot tasks and 5 for 5-shot tasks.

\section{Experiments}
In this section, we conduct extensive experiments to validate the proposed FOT method. We first compare FOT with the following methods on fine grained datasets: 1) typical FSL and FG-FSL methods; 2) state-of-the-art FSL methods; 3) typical hallucination based methods. Secondly, the FOT is used as an data augmentation module to boost some typical FSL methods on FG-FSL tasks. Thirdly, we test the generalization capacity of FOT on Mini-Imagenet dataset. Fourthly, an ablation study is conducted to evaluate the effectiveness of each component. Finally, some visualization results are shown to illustrate that generated samples have meaningful semantics.

\subsection{Datasets}
\label{sec:datasets}
In our experiments, we mainly test our approach on three widely used fine-grained datasets, i.e., Cub birds \cite{wah2011caltech}, Stanford dogs \cite{khosla2011novel} and
Stanford cars \cite{krause20133d}. Detailed statistics are summarized in \autoref{table:dataset}. 

\begin{table}[width=.9\linewidth,cols=4,pos=h]
	\caption{The class split for three fine-grained benchmark datasets. $C_{total}$ is the original number of classes in the datasets, $C_{base}$ is the number of base classes, $C_{val}$ is the number of validation classes and $C_{novel}$ is the number of novel classes.}
	\begin{center}
		\begin{tabular}{lccc}
			\hline
			Dataset & Cub birds          & Stanford Dogs   & Stanford Cars        \\ \hline
			$C_{total}$ & 200          & 120   & 196           \\
			$C_{base}$  & 120          & 70   & 130          \\ 
			$C_{val}$  & 30          & 20   & 17        \\
			$C_{novel}$ & 50          & 30   & 49          \\
			\hline
		\end{tabular}
	\end{center}
	\label{table:dataset}
\end{table}
\textbf{Cub birds} contains 200 categories of birds and a total of 20,580 images \cite{wah2011caltech}. Following the evaluation protocol of \cite{li2019revisiting}, we randomly split the dataset into 120 base, 30 validation, and 50 novel classes.

\textbf{Stanford Dogs} contains 120 categories of dogs and a total of 20,580 images \cite{khosla2011novel}. Following the evaluation protocol of \cite{li2019revisiting}, we randomly split the dataset into 70 base, 20 validation, and 30 novel classes.

\textbf{Stanford Cars} contains 196 categories of cars and a total of 16,185 images \cite{krause20133d}. Following the evaluation protocol of \cite{li2019revisiting}, we randomly split the dataset into 130 base, 17 validation, and 49 novel classes.

We also test the generalization capacity of FOT on the general image dataset, Mini-Imagenet.

\textbf{Mini-Imagenet} contains 100 different categories and 600 random samples in each class from ILSVRC-12 dataset \cite{vinyals2016matching}. Following the evaluation protocol of \cite{sung2018learning}, we randomly split the dataset into 64 base, 16 validation, and 20 novel classes.

\subsection{Experimental settings}

For the proposed FOT method, we apply a published pre-trained BASNet model \cite{qin2019basnet} to obtain saliency maps of all samples, and train the generator $G(\epsilon)$ with $1,000$ epochs, $32$ batch size, $50,000$ quadruplets that satisfy the constrains defined in section \ref{sec:posture_transform}. In the fine-tuning stage, the classifier $C(W_{n})$ is trained with both original and generated samples in support set (FOT$^{*}$ additionally requires unlabeled data in query set). Specifically, we set the number of iterations as 100, and simply adopt a strategy that $C(W_{n})$ is only fed with original samples in the first 40 iterations and mixed-up samples in the rest 60 iterations, which boosts the generalization capability of $C(W_{n})$ effectively.

The other settings of FOT are the same as \cite{chen2019closer}. For typical FSL methods \cite{vinyals2016matching,finn2017model,snell2017prototypical,sung2018learning}, we also take the same settings as \cite{chen2019closer} to ensure fairness. For five specialized FG-FSL methods \cite{wei2019piecewise,huang2019compare,li2019distribution,li2019revisiting,li2020bsnet}, we completely keep the original settings to ensure their performances at the best. In fact, it is disadvantageous for FOT to compare with them, because they generally adopt the best hyperparameters to improve performances, whereas we simply use the same settings as Baseline++ \cite{chen2019closer}. For state-of-the-art FSL methods \cite{wang2019revisiting,qiao2019transductive,ziko2020laplacian,wang2020instance}, we use results reported in their original papers or obtained by reproducing the official codes. For three hallucination based FSL methods\cite{hariharan2017low, schwartz2018delta, wang2018low}, we adopt ConvNet-4, ResNet-18, and ResNet-34 as different backbones for fair and broad comparison. Our experiments are implemented in PyTorch, and models are trained on the Titan Xp GPU using an Adam optimizer.

\subsection{Main results}

\begin{table*}[width=2\linewidth,cols=4,pos=h]
	\caption{Compared results with tyical FSL and FG-FSL methods on three fine-grained datasets (with the same ConvNet-4 backbone). The mean accuracies of the 5-way 1-shot and 5-shot tasks are evaluated in three independent experiments. For each column, the best is bolder and \textbf{{\color{red}red}}, the second best is bolder and \textbf{{\color{blue}blue}}.}
	\small
	\vspace{-2ex}
	\begin{center}
			\begin{tabular}{lcccccc}
				\hline
				\multirow{2}{*}{Method} & \multicolumn{2}{c}{Cub birds} & \multicolumn{2}{c}{Stanford Dogs} & \multicolumn{2}{c}{Stanford Cars} \\
				& 1-shot        & 5-shot        & 1-shot          & 5-shot          & 1-shot          & 5-shot          \\ \hline
				Baseline\cite{chen2019closer}               & $45.97 \pm 0.74$            & $67.09 \pm 0.70$            & $34.42 \pm 0.59$           & $51.95 \pm 0.64$               & $35.60 \pm 0.65$             & $53.08 \pm 0.75$              \\ 
				MatchingNet\cite{vinyals2016matching}              & $57.78 \pm 0.91$            & $72.44 \pm 0.74$            & $42.88 \pm 0.77$              & $58.03 \pm 0.75$              & $41.26 \pm 0.80$              & $62.77 \pm 0.79$              \\
				ProtoNet\cite{snell2017prototypical}                & $44.53 \pm 0.83$            & $75.28 \pm 0.70$            & $37.32 \pm 0.75$               & $59.09 \pm 0.71$               & $30.46 \pm 0.64$             & $61.89 \pm 0.77$              \\
				MAML\cite{finn2017model}                    & $54.92 \pm 0.95$            & $73.18 \pm 0.77$              & $44.64 \pm 0.89$              & $60.20 \pm 0.80$                & $46.71 \pm 0.89$              & $60.73 \pm 0.85$              \\
				RelationNet\cite{sung2018learning}             & $59.58 \pm 0.94$            & $77.62 \pm 0.67$            & $43.05 \pm 0.86$             & $63.42 \pm 0.76$               & $45.48 \pm 0.88$              & $60.26 \pm 0.85$              \\\hline
				PCM\cite{wei2019piecewise}                    & $42.10 \pm 1.96$            & $62.48 \pm 1.21$            & $28.78 \pm 2.33$              & $46.92 \pm 2.00$                & $29.63 \pm 2.38$              & $52.28 \pm 1.46$              \\
				PABN\cite{huang2019compare}             & $63.56 \pm 0.79$            & $75.23 \pm 0.59$            & $45.64 \pm 0.74$            & $58.97 \pm 0.63$               & $53.39 \pm 0.72$               & $66.56 \pm 0.64$              \\
				CovaMNet\cite{li2019distribution}             & $58.51 \pm 0.94$            & $71.15 \pm 0.80$            & $49.10 \pm 0.76$             & $63.04 \pm 0.65$               & $53.85 \pm 0.86$               & $71.33 \pm 0.62$              \\
				DN4\cite{li2019revisiting}             & $55.60 \pm 0.89$            & $77.64 \pm 0.68$            & $45.41 \pm 0.76$             & $63.51 \pm 0.62$               & {\color{red}\bm{$59.84 \pm 0.80$}}               & {\color{red}\bm{$88.65 \pm 0.44$}}              \\
				BSNet(R\&C)\cite{li2020bsnet}             & {\color{blue}\bm{$65.89 \pm 1.00$}}            & {\color{blue}\bm{$80.99 \pm 0.63$}}            & {\color{red}\bm{$51.06 \pm 0.94$}}             & {\color{red}\bm{$68.60 \pm 0.73$}}               & $54.12 \pm 0.96$               & $73.47 \pm 0.75$              \\\hline
				Baseline++\cite{chen2019closer}              & $61.08 \pm 0.84$            & $79.28 \pm 0.68$            & $42.01 \pm 0.73$               & $62.52 \pm 0.72$              & $46.64 \pm 0.80$              & $65.29 \pm 0.73$              \\ 
				FOT (ours)                & {\color{red}\bm{$67.46 \pm 0.68$}}            & {\color{red}\bm{$83.19 \pm 0.43$}}            & {\color{blue}\bm{$49.32 \pm 0.74$}}               & {\color{blue}\bm{$68.18 \pm 0.69$}}               &{\color{blue}\bm{ $54.55 \pm 0.73$}}              & {\color{blue}\bm{$73.69 \pm 0.65$}}              \\\hline
		\end{tabular}
	\end{center}
	
	\vspace{-2ex}
	\label{table:comparison1}
\end{table*}

\textbf{Comparison with typical FSL and FG-FSL methods.} We compare FOT with four typical FSL methods (MatchingNet \cite{vinyals2016matching}, MAML \cite{finn2017model}, ProtoNet \cite{snell2017prototypical}, RelationNet \cite{sung2018learning}) and five specialized FG-FSL methods (PCM \cite{wei2019piecewise}, PABN \cite{huang2019compare}, CovaMNet \cite{li2019distribution}, DN4 \cite{li2019revisiting}, BSNet\cite{li2020bsnet}) on three fine-grained benchmark datasets. For fair comparison, we show the results with the same ConvNet-4 backbone (complete results with other backbones can not be found). \autoref{table:comparison1} shows the compared results on standard 5-way 1-shot and 5-shot protocols. Based on Baseline++ \cite{chen2019closer}, FOT averagely boosts Baseline++ $7.20\%$ on 1-shot and $5.99\%$ on 5-shot. Compared with Baseline and four typical FSL methods \cite{vinyals2016matching,finn2017model,snell2017prototypical,sung2018learning}, FOT exceeds them significantly. Compared with five FG-FSL methods\cite{wei2019piecewise,huang2019compare,li2019distribution,li2019revisiting}, the results of FOT on the bird dataset are the best, while the results of FOT on the car and dog datasets  \cite{khosla2011novel} are the second best. These compared results validate the high stability and strong generalization capacity of FOT.

\textbf{Comparison with state-of-the-art FSL methods.} Recently, some excellent FSL methods (SimpleShot \cite{wang2019revisiting}, TEAM \cite{qiao2019transductive}, LaplacianShot \cite{ziko2020laplacian}, ICI \cite{wang2020instance}, TIM \cite{boudiaf2020transductive}) also show their results on Cub dataset. We compare FOT with these state-of-the-art methods on Cub dataset. For fair comparison, we show the results with the same ResNet-18 backbone (complete results with other backbones can not be found). \autoref{table:stateoftheart} shows: 1) Transductive inference methods usually achieve better performance than inductive inference methods due to the utilization of unlabeled data. 2) FOT has the best results among inductive inference methods, while FOT$^{*}$ also obtain competitive performance with other transductive inference methods. It must be noted that FOT and FOT$^{*}$ only adopt the simplest fine tune architecture. In fact, they have the potential to be applied to other more advanced architectures to achieve better performance.
\begin{table}[h]
	\caption{Compared results with state-of-the-art FSL methods on Cub dataset (with the same ResNet-18 backbone). $(\cdot)^{1}$ are reported in original papers. $(\cdot)^{2}$ are reproduced with the official codes. Note that ICI \cite{wang2020instance} uses the bounding box information to crop the images, but we use the original images for fairness when reproducing. In. and Tran. indicate inductive and transductive setting, respectively. The best results are bolder.}
\begin{tabular}{llll}
\hline
\multirow{2}{*}{Setting} & \multirow{2}{*}{Method} & \multicolumn{2}{c}{Cub birds} \\
                         &                         & 1-shot          & 5-shot         \\ \hline
\multirow{5}{*}{In.}     & Baseline\cite{chen2019closer}                & $65.51^{1}$          & $82.85^{1}$         \\
                         & $\triangle$-encoder\cite{schwartz2018delta}                & $69.81^{1}$          & $84.54^{1}$         \\
                         & SimpleShot\cite{wang2019revisiting}              & $68.90^{1}$         & $84.01^{1}$         \\
                         & Baseline++\cite{chen2019closer}              & $67.02^{2}$          & $83.58^{2}$         \\
                         & FOT (ours)             & \bm{$72.56$}          & \bm{$87.22$}         \\ \hline
\multirow{5}{*}{Trans.}  & TEAM\cite{qiao2019transductive}                & $80.16^{1}$          & $87.17^{1}$         \\
                         & LaplacianShot\cite{ziko2020laplacian}           & $81.00^{1}$           & $88.70^{1}$          \\
                         & TIM-GD\cite{boudiaf2020transductive}                  & $78.72^{2}$          & $87.74^{2}$         \\
                         & ICI\cite{wang2020instance}                    & \bm{$81.34^{2}$}          & $88.32^{2}$         \\
                         & FOT$^{*}$ (ours)              & $80.40^{ }$          & \bm{$89.68^{ }$}         \\ \hline
\end{tabular}
	\label{table:stateoftheart}
\end{table}

\textbf{Comparison with other hallucination based methods.} FOT belongs to hallucination based methods. So we also compare FOT with three hallucination based methods (Linear-hallucinator \cite{hariharan2017low}\footnote{\cite{hariharan2017low} uses a linear offest in the feature space to represent a deformation. For simplicity, we refer to it as Linear-hallucinator in the experiment part.}, Meta-hallucinator \cite{wang2018low}\footnote{\cite{wang2018low} combines meta-learning with hallucination. For simplicity, we refer to it as Meta-hallucinator in the experiment part.}, $\triangle$-encoder \cite{schwartz2018delta}) On Cub dataset. \autoref{table:comparison4} shows the compared results on Cub dataset with three different backbones. The results indicate: 1) Linear-hallucinator \cite{hariharan2017low} degrades the performance of Baseline++ \cite{chen2019closer}. In fact, we find that this method can not converge and obtain effective additional samples on small fine-grained datasets. 2) The performance improvement of the FOT method for Baseline++ is much more significant than that of Meta-hallucinator \cite{wang2018low} or $\triangle$-encoder \cite{schwartz2018delta}. It means that a simple but clear deformation in posture may be more effective than diversified but ambiguous deformations.

\begin{table*}[width=2\linewidth,cols=4,pos=h]
	\caption{Compared results with other hallucination based methods on Cub dataset (with three different backbones). The mean accuracies of the 5-way 1-shot and 5-shot tasks are evaluated in three independent experiments. For each column, the best is bolder and \textbf{{\color{red}red}}, the second best is bolder and \textbf{{\color{blue}blue}}.}
	\vspace{-2ex}
	\begin{center}
		\resizebox{\textwidth}{!}{
			\begin{tabular}{lcccccc}
				\hline
				\multirow{2}{*}{Method} & \multicolumn{2}{c}{ConvNet-4} & \multicolumn{2}{c}{ResNet-18} & \multicolumn{2}{c}{ResNet-34}  \\
				& 1-shot        & 5-shot        & 1-shot          & 5-shot        & 1-shot          & 5-shot             \\ \hline
				Baseline++\cite{chen2019closer}              & $61.08 \pm 0.79$            & $79.28 \pm 0.68$            & $67.02 \pm 0.90$               & $83.58 \pm 0.54$              & $68.00 \pm 0.83$               & $84.50 \pm 0.51$               \\
				Baseline++\cite{chen2019closer} +  Linear-hallucinator\cite{hariharan2017low}           & $59.74 \pm 0.39$            & $78.12 \pm 0.85$            & $64.45 \pm 0.53$               & $81.75 \pm 0.66$              & $64.94 \pm 0.56$               & $82.36 \pm 0.88$                \\ 			Baseline++\cite{chen2019closer} + Meta-hallucinator\cite{wang2018low}              & $63.77 \pm 0.90$            & $80.62 \pm 0.71$            & $69.46 \pm 0.55$               & {\color{blue}\bm{$84.78 \pm 0.78$}}             & $69.33 \pm 0.65$               & $84.87 \pm 0.59$                \\ 		Baseline++\cite{chen2019closer} + $\triangle$-encoder\cite{schwartz2018delta}             & {\color{blue}\bm{$64.47 \pm 0.66$}}            & {\color{blue}\bm{$81.09 \pm 0.68$}}            & {\color{blue}\bm{$69.81 \pm 0.57$}}               & $84.54 \pm 0.92$              & {\color{blue}\bm{$70.63 \pm 0.76$}}               & {\color{blue}\bm{$85.64 \pm 0.37$}}                \\ 		
				Baseline++\cite{chen2019closer} + FOT (ours)                & {\color{red}\bm{$67.46 \pm 0.68$}}            & {\color{red}\bm{$83.19 \pm 0.43$}}            & {\color{red}\bm{$72.56 \pm 0.77$}}               & {\color{red}\bm{$87.22 \pm 0.46$}}           & {\color{red}\bm{$73.38 \pm 0.85$}}               & {\color{red}\bm{$89.01 \pm 0.64$}}                            \\\hline
		\end{tabular}}
		\end{center}
		\vspace{-2ex}
	\label{table:comparison4}
\end{table*}

\textbf{Application results as an auxiliary module.} As a data augmentation method, FOT also can be conveniently applied to any existing FSL methods as an auxiliary module. Specifically, we use the foreground object extractor to process the base class, support and query sets, and then use the posture transformation generator to augment the support set. We conduct experiments by combining FOT with four typical inductive inference FSL methods (with the same ConvNet-4 backbone) and two latest excellent transductive inference FSL methods (with the same ResNet-18 backbone) on Cub dataset. \autoref{table:comparison2} shows that FOT can boost the typical FSL methods significantly. Specifically, MatchingNet \cite{vinyals2016matching} is improved by $5.07\%$ on 1-shot and $5.41\%$ on 5-shot; ProtoNet \cite{snell2017prototypical} is improved by $11.21\%$ on 1-shot and $4.40\%$ on 5-shot; MAML \cite{finn2017model} is improved by $5.10\%$ on 1-shot and $5.76\%$ on 5-shot; RelationNet \cite{sung2018learning} is improved by $5.70\%$ on 1-shot and $4.05\%$ on 5-shot. Surprisingly, the two excellent FSL methods can also be improved effectively. Specifically, TIM-GD \cite{boudiaf2020transductive} is improved by $4.26\%$ on 1-shot and $1.69\%$ on 5-shot; ICI \cite{wang2020instance} is improved by $2.78\%$ on 1-shot and $2.25\%$ on 5-shot. As far as we know, ICI \cite{wang2020instance} and TIM-GD \cite{boudiaf2020transductive} are the state-of-the-art FSL methods. FOT brings them up to the new heights on Cub dataset.

\begin{table}[h]
\caption{Results of combining FOT with typical FSL methods on Cub dataset. In. and Tran. indicate inductive and transductive setting, respectively. The inductive inference methods apdopt the same ConvNet-4 backbone, while the transductive inference methods apdopt the same ResNet-18 backbone. Results of FOT are \textbf{bolder}.}
\begin{tabular}{llll}
\hline
\multirow{2}{*}{Setting} & \multirow{2}{*}{Method} & \multicolumn{2}{c}{Cub birds}   \\
                         &                         & 1-shot         & 5-shot         \\ \hline
\multirow{8}{*}{In.}     & MatchingNet\cite{vinyals2016matching}             & 57.78          & 72.44          \\
                         & MatchingNet + FOT       & \textbf{62.85} & \textbf{77.85} \\
                         & ProtoNet\cite{snell2017prototypical}                & 44.53          & 75.28          \\
                         & ProtoNet + FOT          & \textbf{55.74} & \textbf{79.68} \\
                         & MAML\cite{finn2017model}                    & 54.92          & 73.18          \\
                         & MAML + FOT              & \textbf{60.02} & \textbf{78.94} \\
                         & RelationNet\cite{sung2018learning}            & 59.58          & 77.62          \\
                         & Relationnet + FOT       & \textbf{65.28} & \textbf{81.67} \\ \hline
\multirow{4}{*}{Trans.}   & TIM-GD\cite{boudiaf2020transductive}                  & 78.72          & 87.74          \\
                         & TIM-GD + FOT            & \textbf{82.98} & \textbf{89.43} \\ 
 & ICI\cite{wang2020instance}                     & 81.34          & 88.32          \\
                         & ICI + FOT               & \textbf{84.12} & \textbf{90.57} \\\hline
\end{tabular}

	\vspace{-2ex}
	\label{table:comparison2}
\end{table}

\textbf{Generalization on general FSL tasks.} We test the generalization capacity of FOT on the general image dataset, Mini-Imagenet. \autoref{table:comparison3} shows that FOT can also obtain better performance than the typical FSL methods. Specifically, compared with Baseline++, FOT averagely boosts Baseline++ $3.22\%$ on 1-shot and $1.93\%$ on 5-shot. We observe that results of FOT on Mini-Imagenet are not as significant as those on fine-grained datasets. The main reasons may be: 1) Mini-Imagenet have more complex or multi-objective images, which makes it difficult to extract their foreground objects accurately by the SOD models; 2) 
foreground objects of some categories have no obvious posture characteristics, such as balls and crabs, which degrades the effectiveness of the posture transformation generator. Overall, FOT is more suitable for datasets with single object images and categories possessing rich posture features.

\begin{table}
	\caption{Compared results on Mini-Imagenet dataset (with the same ConvNet-4 backbone). The mean accuracies of the 5-way 1-shot and 5-shot tasks are evaluated in three independent experiments. The results of our FOT method are \textbf{bolder}.}
	\begin{center}
		\begin{tabular}{lcc}
			\hline
			\multirow{2}{*}{Method} & \multicolumn{2}{c}{Mini-Imagenet} \\
			& 1-shot          & 5-shot          \\ \hline
			Baseline \cite{chen2019closer}                & $42.11 \pm 0.71$              & $62.53 \pm 0.69$              \\
			MatchingNet\cite{vinyals2016matching}             & $48.14 \pm 0.78$              & $63.48 \pm 0.66$              \\
			ProtoNet\cite{snell2017prototypical}                & $44.42 \pm 0.84$              & $64.24 \pm 0.72$              \\
			MAML\cite{finn2017model}                    & $46.47 \pm 0.82$              & $62.71 \pm 0.71$              \\
			RelationNet\cite{sung2018learning}             & $49.31 \pm 0.85$              & $66.60 \pm 0.69$              \\ \hline
			Baseline++\cite{chen2019closer}              & $48.24 \pm 0.75$              & $66.43 \pm 0.63$              \\ 
			FOT (ours)                   & \bm{$51.46 \pm 0.76$}              & \bm{$68.36 \pm 0.45$}              \\ \hline
		\end{tabular}
	\end{center}
	\vspace{-2ex}
	
	\label{table:comparison3}
\end{table}

\subsection{Ablation study}
\begin{table*}[width=2\linewidth,cols=4,pos=h]
	\caption{Ablation study on three fine-grained datasets. RB means only removing background of samples, RF means using a bounding box to crop the image and then zoom in the foreground object. The results of integrated FOT method is bolder.}
	\begin{center}
			\begin{tabular}{lcccccc}
				\hline
				\multirow{2}{*}{Method} & \multicolumn{2}{c}{Cub birds} & \multicolumn{2}{c}{Stanford Dogs} & \multicolumn{2}{c}{Stanford Cars} \\
				& 1-shot        & 5-shot        & 1-shot          & 5-shot          & 1-shot          & 5-shot          \\ \hline
				Baseline++\cite{chen2019closer}              & $61.08 \pm 0.84$            & $79.28 \pm 0.68$            & $42.01 \pm 0.73$               & $62.52 \pm 0.72$              & $46.64 \pm 0.80$              & $65.29 \pm 0.73$              \\ \hline
				Baseline++ + RB               & $63.47 \pm 0.85$            & $80.29 \pm 0.59$            & $44.88 \pm 0.77$              & $64.63 \pm 0.75$              & $50.26 \pm 0.80$              & $68.77 \pm 0.79$              \\
				Baseline++ + RB\&RF                & $65.19 \pm 0.89$            & $82.57 \pm 0.60$            & $46.18 \pm 0.75$               & $67.09 \pm 0.71$               & $52.46 \pm 0.64$             & $71.89 \pm 0.77$              \\
				Baseline++ + FOT                      & {\bm{$67.46 \pm 0.68$}}            & {\bm{$83.19 \pm 0.43$}}            & {\bm{$49.32 \pm 0.74$}}               & {\bm{$68.18 \pm 0.69$}}               & {\bm{$54.55 \pm 0.73$}}              & {\bm{$73.69 \pm 0.65$}}            \\ \hline
		\end{tabular}
	\end{center}
	\vspace{-2ex}	
	\label{table:ablation}
\end{table*}
To get a better understanding of different components in FOT, we conduct the ablation study. In specific, we split the foreground object extractor into two stages: removing the background (for example, the fourth image in \autoref{fig:s1}) and resizing the foreground (for example, the sixth image in \autoref{fig:s1}). We take the posture transformation generator as a single component. \autoref{table:ablation} shows the ablation results on three benchmark datasets. Compared with Baseline++ method, simply removing background can averagely increase the accuracy by $2.96\%$ on 1-shot and $2.20\%$ on 5-shot. It validates that image background tends to play a negative role in FG-FSL tasks. Resizing the foreground object increases the accuracy by $1.74\%$ on 1-shot and $2.62\%$ on 5-shot. It verifies that a zoomed-in foreground object is more favorable for FG-FSL tasks. Additional samples generated by the posture transformation generator increases the accuracy by $2.50\%$ on 1-shot and $1.17\%$ on 5-shot, which means that it is an effective data augmentation method by transforming the posture of foreground objects.

\subsection{Visualization results}
\begin{figure*}
	\begin{center}
		\includegraphics[width=0.9\linewidth]{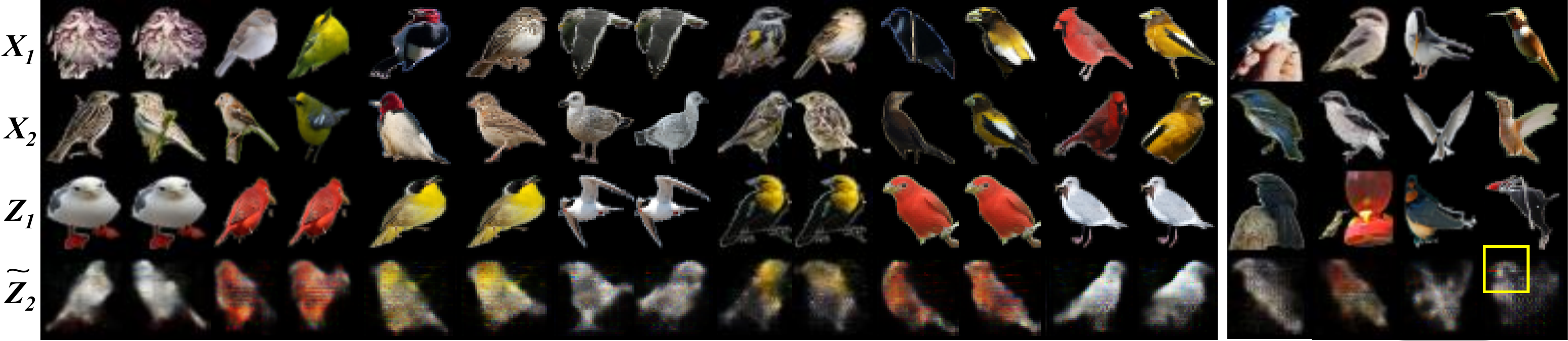}
	\end{center}
	\vspace{-2ex}
	\caption{Visualization results of generated samples. Each column represents a quadruplet $\{X_{1}, X_{2}, Z_{1}, \widetilde{Z_{2}}\}$. $(X_{1}, X_{2})$ represents a pair of samples from a base class. $Z_{1}$ represents a sample from a novel class, which has similar posture with $X_{1}$. $\widetilde{Z_{2}}$ represents the generated sample. The left side shows some well generated samples, and the right side shows some bad ones. }
	\label{fig:Visualization1}
\end{figure*}
To demonstrate that FOT is able to generate meaningful semantically augmented samples, we show some generated images in \autoref{fig:Visualization1}. Each column represents a quadruplet $\{X_{1}, X_{2}, Z_{1}, \widetilde{Z_{2}}\}$. $(X_{1}, X_{2})$ represents a pair of samples from a base class. $Z_{1}$ represents a sample from a novel class, which has similar posture with $X_{1}$. $\widetilde{Z_{2}}$ represents the generated sample. The left side shows some well generated samples. Obviously, each generated sample $\widetilde{Z_{2}}$ have the posture of $X_{2}$ and the class feature of $Z_{1}$. This means that the generator works well. The right side shows some bad cases. We find that the bad samples are mainly caused by two factors. One is that the foreground object extractor does not capture the accurate foreground object. The other is that it is difficult to learn a rare posture transformation while there are not enough similar training quadruplets in the training dataset $D_{g}$. 

However, some bad generated samples also contain some features of samples in the novel class, which can still play a positive role in FG-FSL tasks. The rightmost column of \autoref{fig:Visualization1} shows a typical example. The generated bird does not look like a flying bird, but the red beak (framed by the yellow rectangle) still indicates its true category.

\section{Conclusion}

In this paper, we have proposed a novel data augmentation method to deal with FG-FSL tasks, named foreground object transformation (FOT). It mainly consists of two carefully designed components, a foreground object extractor and a posture transformation generator. Essentially, the former decreases intra-class variance by removing the image background, while the latter increases sample diversity by strengthening the features of foreground objects and generating additional samples with different postures. Experimental results have validated that our method can boost simple fine-tuning baselines to a competitive level with the state-of-the-art methods both in inductive setting and transductive setting. It also brings the latest FSL methods up to the new state-of-the-art on FG-FSL tasks. Moreover, FOT can also be easily extended to more general image classification tasks. Currently, the performance gain of our method may be less significant when dealing with complex tasks, such as datasets with multi-object images. In the future, extending FOT to handle more complicated images can be considered as an interesting research direction.

\section*{Acknowledgment}

This work is supported in part by the National Science and Technology Major Project of the Ministry of Science and Technology of China under Grants 2018AAA0101604, the National Natural Science Foundation of China under Grants 61906106, 62022048, and 61803321.

\bibliographystyle{cas-model2-names}

\bibliography{cas-refs}

\begin{thebibliography}{71}
\expandafter\ifx\csname natexlab\endcsname\relax\def\natexlab#1{#1}\fi
\providecommand{\url}[1]{\texttt{#1}}
\providecommand{\href}[2]{#2}
\providecommand{\path}[1]{#1}
\providecommand{\DOIprefix}{doi:}
\providecommand{\ArXivprefix}{arXiv:}
\providecommand{\URLprefix}{URL: }
\providecommand{\Pubmedprefix}{pmid:}
\providecommand{\doi}[1]{\href{http://dx.doi.org/#1}{\path{#1}}}
\providecommand{\Pubmed}[1]{\href{pmid:#1}{\path{#1}}}
\providecommand{\bibinfo}[2]{#2}
\ifx\xfnm\relax \def\xfnm[#1]{\unskip,\space#1}\fi
\bibitem[{Boudiaf et~al.(2020)Boudiaf, Masud, Rony, Dolz, Piantanida and
  Ayed}]{boudiaf2020transductive}
\bibinfo{author}{Boudiaf, M.}, \bibinfo{author}{Masud, Z.I.},
  \bibinfo{author}{Rony, J.}, \bibinfo{author}{Dolz, J.},
  \bibinfo{author}{Piantanida, P.}, \bibinfo{author}{Ayed, I.B.},
  \bibinfo{year}{2020}.
\newblock \bibinfo{title}{Transductive information maximization for few-shot
  learning}.
\newblock \bibinfo{journal}{arXiv preprint arXiv:2008.11297} .
\bibitem[{Chen et~al.(2018)Chen, Tan, Wang and Hu}]{chen2018reverse}
\bibinfo{author}{Chen, S.}, \bibinfo{author}{Tan, X.}, \bibinfo{author}{Wang,
  B.}, \bibinfo{author}{Hu, X.}, \bibinfo{year}{2018}.
\newblock \bibinfo{title}{Reverse attention for salient object detection}, in:
  \bibinfo{booktitle}{Proceedings of the European Conference on Computer Vision
  (ECCV)}, pp. \bibinfo{pages}{234--250}.
\bibitem[{Chen et~al.(2019a)Chen, Liu, KiraAuthors, Chiang and
  Jiabin}]{chen2019closer}
\bibinfo{author}{Chen, W.}, \bibinfo{author}{Liu, Y.},
  \bibinfo{author}{KiraAuthors, Z.}, \bibinfo{author}{Chiang, Y.},
  \bibinfo{author}{Jiabin, H.}, \bibinfo{year}{2019}a.
\newblock \bibinfo{title}{A closer look at few-shot classification}, in:
  \bibinfo{booktitle}{Proceedings of the IEEE International Conference on
  Learning Representations Worshops}.
\bibitem[{Chen et~al.(2019b)Chen, Fu, Wang, Ma, Liu and Hebert}]{chen2019image}
\bibinfo{author}{Chen, Z.}, \bibinfo{author}{Fu, Y.}, \bibinfo{author}{Wang,
  Y.X.}, \bibinfo{author}{Ma, L.}, \bibinfo{author}{Liu, W.},
  \bibinfo{author}{Hebert, M.}, \bibinfo{year}{2019}b.
\newblock \bibinfo{title}{Image deformation meta-networks for one-shot
  learning}, in: \bibinfo{booktitle}{Proceedings of the IEEE Conference on
  Computer Vision and Pattern Recognition}, pp. \bibinfo{pages}{8680--8689}.
\bibitem[{Chen et~al.(2019c)Chen, Fu, Zhang, Jiang, Xue and
  Sigal}]{chen2019multi}
\bibinfo{author}{Chen, Z.}, \bibinfo{author}{Fu, Y.}, \bibinfo{author}{Zhang,
  Y.}, \bibinfo{author}{Jiang, Y.G.}, \bibinfo{author}{Xue, X.},
  \bibinfo{author}{Sigal, L.}, \bibinfo{year}{2019}c.
\newblock \bibinfo{title}{Multi-level semantic feature augmentation for
  one-shot learning}.
\newblock \bibinfo{journal}{IEEE Transactions on Image Processing}
  \bibinfo{volume}{28}, \bibinfo{pages}{4594--4605}.
\bibitem[{Choe et~al.(2020)Choe, Lee and Shim}]{choe2020attention}
\bibinfo{author}{Choe, J.}, \bibinfo{author}{Lee, S.}, \bibinfo{author}{Shim,
  H.}, \bibinfo{year}{2020}.
\newblock \bibinfo{title}{Attention-based dropout layer for weakly supervised
  single object localization and semantic segmentation}.
\newblock \bibinfo{journal}{IEEE transactions on pattern analysis and machine
  intelligence} .
\bibitem[{Dhillon et~al.(2019)Dhillon, Chaudhari, Ravichandran and
  Soatto}]{dhillon2019baseline}
\bibinfo{author}{Dhillon, G.S.}, \bibinfo{author}{Chaudhari, P.},
  \bibinfo{author}{Ravichandran, A.}, \bibinfo{author}{Soatto, S.},
  \bibinfo{year}{2019}.
\newblock \bibinfo{title}{A baseline for few-shot image classification}.
\newblock \bibinfo{journal}{arXiv preprint arXiv:1909.02729} .
\bibitem[{Finn et~al.(2017)Finn, Abbeel and Levine}]{finn2017model}
\bibinfo{author}{Finn, C.}, \bibinfo{author}{Abbeel, P.},
  \bibinfo{author}{Levine, S.}, \bibinfo{year}{2017}.
\newblock \bibinfo{title}{Model-agnostic meta-learning for fast adaptation of
  deep networks}, in: \bibinfo{booktitle}{Proceedings of the 34th International
  Conference on Machine Learning-Volume 70}, \bibinfo{organization}{JMLR. org}.
  pp. \bibinfo{pages}{1126--1135}.
\bibitem[{Finn et~al.(2018)Finn, Xu and Levine}]{finn2018probabilistic}
\bibinfo{author}{Finn, C.}, \bibinfo{author}{Xu, K.}, \bibinfo{author}{Levine,
  S.}, \bibinfo{year}{2018}.
\newblock \bibinfo{title}{Probabilistic model-agnostic meta-learning}, in:
  \bibinfo{booktitle}{Advances in Neural Information Processing Systems}, pp.
  \bibinfo{pages}{9516--9527}.
\bibitem[{Fu et~al.(2017)Fu, Zheng and Mei}]{fu2017look}
\bibinfo{author}{Fu, J.}, \bibinfo{author}{Zheng, H.}, \bibinfo{author}{Mei,
  T.}, \bibinfo{year}{2017}.
\newblock \bibinfo{title}{Look closer to see better: Recurrent attention
  convolutional neural network for fine-grained image recognition}, in:
  \bibinfo{booktitle}{Proceedings of the IEEE Conference on Computer Vision and
  Pattern Recognition}, pp. \bibinfo{pages}{4438--4446}.
\bibitem[{Gao et~al.(2018)Gao, Shou, Zareian, Zhang and Chang}]{gao2018low}
\bibinfo{author}{Gao, H.}, \bibinfo{author}{Shou, Z.},
  \bibinfo{author}{Zareian, A.}, \bibinfo{author}{Zhang, H.},
  \bibinfo{author}{Chang, S.F.}, \bibinfo{year}{2018}.
\newblock \bibinfo{title}{Low-shot learning via covariance-preserving
  adversarial augmentation networks}.
\newblock \bibinfo{journal}{arXiv preprint arXiv:1810.11730} .
\bibitem[{Garcia and Bruna(2017)}]{garcia2017few}
\bibinfo{author}{Garcia, V.}, \bibinfo{author}{Bruna, J.},
  \bibinfo{year}{2017}.
\newblock \bibinfo{title}{Few-shot learning with graph neural networks}.
\newblock \bibinfo{journal}{arXiv preprint arXiv:1711.04043} .
\bibitem[{Ge et~al.(2019)Ge, Lin and Yu}]{ge2019weakly}
\bibinfo{author}{Ge, W.}, \bibinfo{author}{Lin, X.}, \bibinfo{author}{Yu, Y.},
  \bibinfo{year}{2019}.
\newblock \bibinfo{title}{Weakly supervised complementary parts models for
  fine-grained image classification from the bottom up}, in:
  \bibinfo{booktitle}{Proceedings of the IEEE Conference on Computer Vision and
  Pattern Recognition}, pp. \bibinfo{pages}{3034--3043}.
\bibitem[{Hariharan and Girshick(2017)}]{hariharan2017low}
\bibinfo{author}{Hariharan, B.}, \bibinfo{author}{Girshick, R.},
  \bibinfo{year}{2017}.
\newblock \bibinfo{title}{Low-shot visual recognition by shrinking and
  hallucinating features}, in: \bibinfo{booktitle}{Proceedings of the IEEE
  International Conference on Computer Vision}, pp.
  \bibinfo{pages}{3018--3027}.
\bibitem[{He et~al.(2016)He, Zhang, Ren and Sun}]{he2016deep}
\bibinfo{author}{He, K.}, \bibinfo{author}{Zhang, X.}, \bibinfo{author}{Ren,
  S.}, \bibinfo{author}{Sun, J.}, \bibinfo{year}{2016}.
\newblock \bibinfo{title}{Deep residual learning for image recognition}, in:
  \bibinfo{booktitle}{Proceedings of the IEEE Conference on Computer Vision and
  Pattern Recognition}, pp. \bibinfo{pages}{770--778}.
\bibitem[{He et~al.(2018)He, Peng and Zhao}]{he2018fast}
\bibinfo{author}{He, X.}, \bibinfo{author}{Peng, Y.}, \bibinfo{author}{Zhao,
  J.}, \bibinfo{year}{2018}.
\newblock \bibinfo{title}{Fast fine-grained image classification via weakly
  supervised discriminative localization}.
\newblock \bibinfo{journal}{IEEE Transactions on Circuits and Systems for Video
  Technology} \bibinfo{volume}{29}, \bibinfo{pages}{1394--1407}.
\bibitem[{Hou et~al.(2019)Hou, Chang, Ma, Shan and Chen}]{hou2019cross}
\bibinfo{author}{Hou, R.}, \bibinfo{author}{Chang, H.}, \bibinfo{author}{Ma,
  B.}, \bibinfo{author}{Shan, S.}, \bibinfo{author}{Chen, X.},
  \bibinfo{year}{2019}.
\newblock \bibinfo{title}{Cross attention network for few-shot classification}.
\newblock \bibinfo{journal}{arXiv preprint arXiv:1910.07677} .
\bibitem[{Hu and Qi(2019)}]{hu2019see}
\bibinfo{author}{Hu, T.}, \bibinfo{author}{Qi, H.}, \bibinfo{year}{2019}.
\newblock \bibinfo{title}{See better before looking closer: Weakly supervised
  data augmentation network for fine-grained visual classification}.
\newblock \bibinfo{journal}{arXiv preprint arXiv:1901.09891} .
\bibitem[{Hu et~al.(2018)Hu, Xu, Huang, Qi, Huang and Lu}]{hu2018weakly}
\bibinfo{author}{Hu, T.}, \bibinfo{author}{Xu, J.}, \bibinfo{author}{Huang,
  C.}, \bibinfo{author}{Qi, H.}, \bibinfo{author}{Huang, Q.},
  \bibinfo{author}{Lu, Y.}, \bibinfo{year}{2018}.
\newblock \bibinfo{title}{Weakly supervised bilinear attention network for
  fine-grained visual classification}.
\newblock \bibinfo{journal}{arXiv preprint arXiv:1808.02152} .
\bibitem[{Huang et~al.(2019)Huang, Zhang, Zhang, Wu and Xu}]{huang2019compare}
\bibinfo{author}{Huang, H.}, \bibinfo{author}{Zhang, J.},
  \bibinfo{author}{Zhang, J.}, \bibinfo{author}{Wu, Q.}, \bibinfo{author}{Xu,
  J.}, \bibinfo{year}{2019}.
\newblock \bibinfo{title}{Compare more nuanced: Pairwise alignment bilinear
  network for few-shot fine-grained learning}.
\newblock \bibinfo{journal}{arXiv preprint arXiv:1904.03580} .
\bibitem[{Huang et~al.(2016)Huang, Xu, Tao and Zhang}]{huang2016part}
\bibinfo{author}{Huang, S.}, \bibinfo{author}{Xu, Z.}, \bibinfo{author}{Tao,
  D.}, \bibinfo{author}{Zhang, Y.}, \bibinfo{year}{2016}.
\newblock \bibinfo{title}{Part-stacked cnn for fine-grained visual
  categorization}, in: \bibinfo{booktitle}{Proceedings of the IEEE Conference
  on Computer Vision and Pattern Recognition}, pp. \bibinfo{pages}{1173--1182}.
\bibitem[{Ji et~al.(2019)Ji, Wen, Zhang, Du, Wu, Zhao, Liu and
  Huang}]{ji2019attention}
\bibinfo{author}{Ji, R.}, \bibinfo{author}{Wen, L.}, \bibinfo{author}{Zhang,
  L.}, \bibinfo{author}{Du, D.}, \bibinfo{author}{Wu, Y.},
  \bibinfo{author}{Zhao, C.}, \bibinfo{author}{Liu, X.},
  \bibinfo{author}{Huang, F.}, \bibinfo{year}{2019}.
\newblock \bibinfo{title}{Attention convolutional binary neural tree for
  fine-grained visual categorization}.
\newblock \bibinfo{journal}{arXiv preprint arXiv:1909.11378} .
\bibitem[{Jiang et~al.(2020)Jiang, Kang, Zhou and Feng}]{jiang2020few}
\bibinfo{author}{Jiang, Z.}, \bibinfo{author}{Kang, B.}, \bibinfo{author}{Zhou,
  K.}, \bibinfo{author}{Feng, J.}, \bibinfo{year}{2020}.
\newblock \bibinfo{title}{Few-shot classification via adaptive attention}.
\newblock \bibinfo{journal}{arXiv preprint arXiv:2008.02465} .
\bibitem[{Khosla et~al.(2011)Khosla, Jayadevaprakash, Yao and
  Li}]{khosla2011novel}
\bibinfo{author}{Khosla, A.}, \bibinfo{author}{Jayadevaprakash, N.},
  \bibinfo{author}{Yao, B.}, \bibinfo{author}{Li, F.F.}, \bibinfo{year}{2011}.
\newblock \bibinfo{title}{Novel dataset for fine-grained image categorization:
  Stanford dogs}, in: \bibinfo{booktitle}{Proc. CVPR Workshop on Fine-Grained
  Visual Categorization (FGVC)}.
\bibitem[{Krause et~al.(2013)Krause, Stark, Deng and Fei-Fei}]{krause20133d}
\bibinfo{author}{Krause, J.}, \bibinfo{author}{Stark, M.},
  \bibinfo{author}{Deng, J.}, \bibinfo{author}{Fei-Fei, L.},
  \bibinfo{year}{2013}.
\newblock \bibinfo{title}{3d object representations for fine-grained
  categorization}, in: \bibinfo{booktitle}{Proceedings of the IEEE
  International Conference on Computer Vision Workshops}, pp.
  \bibinfo{pages}{554--561}.
\bibitem[{Li and Yu(2015)}]{li2015visual}
\bibinfo{author}{Li, G.}, \bibinfo{author}{Yu, Y.}, \bibinfo{year}{2015}.
\newblock \bibinfo{title}{Visual saliency based on multiscale deep features},
  in: \bibinfo{booktitle}{Proceedings of the IEEE Conference on Computer Vision
  and Pattern Recognition}, pp. \bibinfo{pages}{5455--5463}.
\bibitem[{Li et~al.(2019a)Li, Wang, Xu, Huo, Gao and Luo}]{li2019revisiting}
\bibinfo{author}{Li, W.}, \bibinfo{author}{Wang, L.}, \bibinfo{author}{Xu, J.},
  \bibinfo{author}{Huo, J.}, \bibinfo{author}{Gao, Y.}, \bibinfo{author}{Luo,
  J.}, \bibinfo{year}{2019}a.
\newblock \bibinfo{title}{Revisiting local descriptor based image-to-class
  measure for few-shot learning}, in: \bibinfo{booktitle}{Proceedings of the
  IEEE Conference on Computer Vision and Pattern Recognition}, pp.
  \bibinfo{pages}{7260--7268}.
\bibitem[{Li et~al.(2019b)Li, Xu, Huo, Wang, Gao and Luo}]{li2019distribution}
\bibinfo{author}{Li, W.}, \bibinfo{author}{Xu, J.}, \bibinfo{author}{Huo, J.},
  \bibinfo{author}{Wang, L.}, \bibinfo{author}{Gao, Y.}, \bibinfo{author}{Luo,
  J.}, \bibinfo{year}{2019}b.
\newblock \bibinfo{title}{Distribution consistency based covariance metric
  networks for few-shot learning}, in: \bibinfo{booktitle}{Proceedings of the
  AAAI Conference on Artificial Intelligence}, pp. \bibinfo{pages}{8642--8649}.
\bibitem[{Li et~al.(2019c)Li, Sun, Liu, Zhou, Zheng, Chua and
  Schiele}]{li2019learning}
\bibinfo{author}{Li, X.}, \bibinfo{author}{Sun, Q.}, \bibinfo{author}{Liu, Y.},
  \bibinfo{author}{Zhou, Q.}, \bibinfo{author}{Zheng, S.},
  \bibinfo{author}{Chua, T.S.}, \bibinfo{author}{Schiele, B.},
  \bibinfo{year}{2019}c.
\newblock \bibinfo{title}{Learning to self-train for semi-supervised few-shot
  classification}.
\newblock \bibinfo{journal}{Advances in Neural Information Processing Systems}
  \bibinfo{volume}{32}, \bibinfo{pages}{10276--10286}.
\bibitem[{Li et~al.(2020)Li, Wu, Sun, Ma, Cao and Xue}]{li2020bsnet}
\bibinfo{author}{Li, X.}, \bibinfo{author}{Wu, J.}, \bibinfo{author}{Sun, Z.},
  \bibinfo{author}{Ma, Z.}, \bibinfo{author}{Cao, J.}, \bibinfo{author}{Xue,
  J.H.}, \bibinfo{year}{2020}.
\newblock \bibinfo{title}{Bsnet: Bi-similarity network for few-shot
  fine-grained image classification}.
\newblock \bibinfo{journal}{IEEE Transactions on Image Processing}
  \bibinfo{volume}{30}, \bibinfo{pages}{1318--1331}.
\bibitem[{Li et~al.(2018)Li, Yang, Cheng, Liu and Shen}]{li2018contour}
\bibinfo{author}{Li, X.}, \bibinfo{author}{Yang, F.}, \bibinfo{author}{Cheng,
  H.}, \bibinfo{author}{Liu, W.}, \bibinfo{author}{Shen, D.},
  \bibinfo{year}{2018}.
\newblock \bibinfo{title}{Contour knowledge transfer for salient object
  detection}, in: \bibinfo{booktitle}{Proceedings of the European Conference on
  Computer Vision (ECCV)}, pp. \bibinfo{pages}{355--370}.
\bibitem[{Lim et~al.(2021)Lim, Lim, Ooi and Lee}]{lim2021efficient}
\bibinfo{author}{Lim, J.Y.}, \bibinfo{author}{Lim, K.M.}, \bibinfo{author}{Ooi,
  S.Y.}, \bibinfo{author}{Lee, C.P.}, \bibinfo{year}{2021}.
\newblock \bibinfo{title}{Efficient-prototypicalnet with self knowledge
  distillation for few-shot learning}.
\newblock \bibinfo{journal}{Neurocomputing} \bibinfo{volume}{459},
  \bibinfo{pages}{327--337}.
\bibitem[{Liu et~al.(2019)Liu, Song and Qin}]{liu2019prototype}
\bibinfo{author}{Liu, J.}, \bibinfo{author}{Song, L.}, \bibinfo{author}{Qin,
  Y.}, \bibinfo{year}{2019}.
\newblock \bibinfo{title}{Prototype rectification for few-shot learning}.
\newblock \bibinfo{journal}{arXiv preprint arXiv:1911.10713} .
\bibitem[{Liu et~al.(2018a)Liu, Han and Yang}]{liu2018picanet}
\bibinfo{author}{Liu, N.}, \bibinfo{author}{Han, J.}, \bibinfo{author}{Yang,
  M.H.}, \bibinfo{year}{2018}a.
\newblock \bibinfo{title}{Picanet: Learning pixel-wise contextual attention for
  saliency detection}, in: \bibinfo{booktitle}{Proceedings of the IEEE
  Conference on Computer Vision and Pattern Recognition}, pp.
  \bibinfo{pages}{3089--3098}.
\bibitem[{Liu et~al.(2020)Liu, Zhou, Liu and Jiang}]{liu2020meta}
\bibinfo{author}{Liu, X.}, \bibinfo{author}{Zhou, F.}, \bibinfo{author}{Liu,
  J.}, \bibinfo{author}{Jiang, L.}, \bibinfo{year}{2020}.
\newblock \bibinfo{title}{Meta-learning based prototype-relation network for
  few-shot classification}.
\newblock \bibinfo{journal}{Neurocomputing} \bibinfo{volume}{383},
  \bibinfo{pages}{224--234}.
\bibitem[{Liu et~al.(2018b)Liu, Lee, Park, Kim, Yang, Hwang and
  Yang}]{liu2018learning}
\bibinfo{author}{Liu, Y.}, \bibinfo{author}{Lee, J.}, \bibinfo{author}{Park,
  M.}, \bibinfo{author}{Kim, S.}, \bibinfo{author}{Yang, E.},
  \bibinfo{author}{Hwang, S.J.}, \bibinfo{author}{Yang, Y.},
  \bibinfo{year}{2018}b.
\newblock \bibinfo{title}{Learning to propagate labels: Transductive
  propagation network for few-shot learning}.
\newblock \bibinfo{journal}{arXiv preprint arXiv:1805.10002} .
\bibitem[{Long et~al.(2015)Long, Shelhamer and Darrell}]{long2015fully}
\bibinfo{author}{Long, J.}, \bibinfo{author}{Shelhamer, E.},
  \bibinfo{author}{Darrell, T.}, \bibinfo{year}{2015}.
\newblock \bibinfo{title}{Fully convolutional networks for semantic
  segmentation}, in: \bibinfo{booktitle}{Proceedings of the IEEE Conference on
  Computer Vision and Pattern Recognition}, pp. \bibinfo{pages}{3431--3440}.
\bibitem[{Nichol and Schulman(2018)}]{nichol2018reptile}
\bibinfo{author}{Nichol, A.}, \bibinfo{author}{Schulman, J.},
  \bibinfo{year}{2018}.
\newblock \bibinfo{title}{Reptile: a scalable metalearning algorithm}.
\newblock \bibinfo{journal}{arXiv preprint arXiv:1803.02999}
  \bibinfo{volume}{2}.
\bibitem[{Qiao et~al.(2019)Qiao, Shi, Li, Wang, Huang and
  Tian}]{qiao2019transductive}
\bibinfo{author}{Qiao, L.}, \bibinfo{author}{Shi, Y.}, \bibinfo{author}{Li,
  J.}, \bibinfo{author}{Wang, Y.}, \bibinfo{author}{Huang, T.},
  \bibinfo{author}{Tian, Y.}, \bibinfo{year}{2019}.
\newblock \bibinfo{title}{Transductive episodic-wise adaptive metric for
  few-shot learning}, in: \bibinfo{booktitle}{Proceedings of the IEEE/CVF
  International Conference on Computer Vision}, pp.
  \bibinfo{pages}{3603--3612}.
\bibitem[{Qin et~al.(2019)Qin, Zhang, Huang, Gao, Dehghan and
  Jagersand}]{qin2019basnet}
\bibinfo{author}{Qin, X.}, \bibinfo{author}{Zhang, Z.}, \bibinfo{author}{Huang,
  C.}, \bibinfo{author}{Gao, C.}, \bibinfo{author}{Dehghan, M.},
  \bibinfo{author}{Jagersand, M.}, \bibinfo{year}{2019}.
\newblock \bibinfo{title}{Basnet: Boundary-aware salient object detection}, in:
  \bibinfo{booktitle}{Proceedings of the IEEE Conference on Computer Vision and
  Pattern Recognition}, pp. \bibinfo{pages}{7479--7489}.
\bibitem[{Ren et~al.(2018)Ren, Triantafillou, Ravi, Snell, Swersky, Tenenbaum,
  Larochelle and Zemel}]{ren2018meta}
\bibinfo{author}{Ren, M.}, \bibinfo{author}{Triantafillou, E.},
  \bibinfo{author}{Ravi, S.}, \bibinfo{author}{Snell, J.},
  \bibinfo{author}{Swersky, K.}, \bibinfo{author}{Tenenbaum, J.B.},
  \bibinfo{author}{Larochelle, H.}, \bibinfo{author}{Zemel, R.S.},
  \bibinfo{year}{2018}.
\newblock \bibinfo{title}{Meta-learning for semi-supervised few-shot
  classification}.
\newblock \bibinfo{journal}{arXiv preprint arXiv:1803.00676} .
\bibitem[{Rusu et~al.(2018)Rusu, Rao, Sygnowski, Vinyals, Pascanu, Osindero and
  Hadsell}]{rusu2018meta}
\bibinfo{author}{Rusu, A.A.}, \bibinfo{author}{Rao, D.},
  \bibinfo{author}{Sygnowski, J.}, \bibinfo{author}{Vinyals, O.},
  \bibinfo{author}{Pascanu, R.}, \bibinfo{author}{Osindero, S.},
  \bibinfo{author}{Hadsell, R.}, \bibinfo{year}{2018}.
\newblock \bibinfo{title}{Meta-learning with latent embedding optimization}.
\newblock \bibinfo{journal}{arXiv preprint arXiv:1807.05960} .
\bibitem[{Schwartz et~al.(2018)Schwartz, Karlinsky, Shtok, Harary, Marder,
  Kumar, Feris, Giryes and Bronstein}]{schwartz2018delta}
\bibinfo{author}{Schwartz, E.}, \bibinfo{author}{Karlinsky, L.},
  \bibinfo{author}{Shtok, J.}, \bibinfo{author}{Harary, S.},
  \bibinfo{author}{Marder, M.}, \bibinfo{author}{Kumar, A.},
  \bibinfo{author}{Feris, R.}, \bibinfo{author}{Giryes, R.},
  \bibinfo{author}{Bronstein, A.}, \bibinfo{year}{2018}.
\newblock \bibinfo{title}{Delta-encoder: an effective sample synthesis method
  for few-shot object recognition}, in: \bibinfo{booktitle}{Advances in Neural
  Information Processing Systems}, pp. \bibinfo{pages}{2845--2855}.
\bibitem[{Sermanet et~al.(2014)Sermanet, Frome and
  Real}]{sermanet2014attention}
\bibinfo{author}{Sermanet, P.}, \bibinfo{author}{Frome, A.},
  \bibinfo{author}{Real, E.}, \bibinfo{year}{2014}.
\newblock \bibinfo{title}{Attention for fine-grained categorization}.
\newblock \bibinfo{journal}{arXiv preprint arXiv:1412.7054} .
\bibitem[{Simonyan and Zisserman(2014)}]{simonyan2014very}
\bibinfo{author}{Simonyan, K.}, \bibinfo{author}{Zisserman, A.},
  \bibinfo{year}{2014}.
\newblock \bibinfo{title}{Very deep convolutional networks for large-scale
  image recognition}.
\newblock \bibinfo{journal}{arXiv preprint arXiv:1409.1556} .
\bibitem[{Snell et~al.(2017)Snell, Swersky and Zemel}]{snell2017prototypical}
\bibinfo{author}{Snell, J.}, \bibinfo{author}{Swersky, K.},
  \bibinfo{author}{Zemel, R.}, \bibinfo{year}{2017}.
\newblock \bibinfo{title}{Prototypical networks for few-shot learning}, in:
  \bibinfo{booktitle}{Advances in Neural Information Processing Systems}, pp.
  \bibinfo{pages}{4077--4087}.
\bibitem[{Sung et~al.(2018)Sung, Yang, Zhang, Xiang, Torr and
  Hospedales}]{sung2018learning}
\bibinfo{author}{Sung, F.}, \bibinfo{author}{Yang, Y.}, \bibinfo{author}{Zhang,
  L.}, \bibinfo{author}{Xiang, T.}, \bibinfo{author}{Torr, P.H.},
  \bibinfo{author}{Hospedales, T.M.}, \bibinfo{year}{2018}.
\newblock \bibinfo{title}{Learning to compare: Relation network for few-shot
  learning}, in: \bibinfo{booktitle}{Proceedings of the IEEE Conference on
  Computer Vision and Pattern Recognition}, pp. \bibinfo{pages}{1199--1208}.
\bibitem[{Vinyals et~al.(2016)Vinyals, Blundell, Lillicrap, Wierstra
  et~al.}]{vinyals2016matching}
\bibinfo{author}{Vinyals, O.}, \bibinfo{author}{Blundell, C.},
  \bibinfo{author}{Lillicrap, T.}, \bibinfo{author}{Wierstra, D.}, et~al.,
  \bibinfo{year}{2016}.
\newblock \bibinfo{title}{Matching networks for one shot learning}, in:
  \bibinfo{booktitle}{Advances in Neural Information Processing Systems}, pp.
  \bibinfo{pages}{3630--3638}.
\bibitem[{Wah et~al.(2011)Wah, Branson, Welinder, Perona and
  Belongie}]{wah2011caltech}
\bibinfo{author}{Wah, C.}, \bibinfo{author}{Branson, S.},
  \bibinfo{author}{Welinder, P.}, \bibinfo{author}{Perona, P.},
  \bibinfo{author}{Belongie, S.}, \bibinfo{year}{2011}.
\newblock \bibinfo{title}{The caltech-ucsd birds-200-2011 dataset} .
\bibitem[{Wang et~al.(2015)Wang, Lu, Ruan and Yang}]{wang2015deep}
\bibinfo{author}{Wang, L.}, \bibinfo{author}{Lu, H.}, \bibinfo{author}{Ruan,
  X.}, \bibinfo{author}{Yang, M.H.}, \bibinfo{year}{2015}.
\newblock \bibinfo{title}{Deep networks for saliency detection via local
  estimation and global search}, in: \bibinfo{booktitle}{Proceedings of the
  IEEE Conference on Computer Vision and Pattern Recognition}, pp.
  \bibinfo{pages}{3183--3192}.
\bibitem[{Wang et~al.(2017)Wang, Lu, Wang, Feng, Wang, Yin and
  Ruan}]{wang2017learning}
\bibinfo{author}{Wang, L.}, \bibinfo{author}{Lu, H.}, \bibinfo{author}{Wang,
  Y.}, \bibinfo{author}{Feng, M.}, \bibinfo{author}{Wang, D.},
  \bibinfo{author}{Yin, B.}, \bibinfo{author}{Ruan, X.}, \bibinfo{year}{2017}.
\newblock \bibinfo{title}{Learning to detect salient objects with image-level
  supervision}, in: \bibinfo{booktitle}{Proceedings of the IEEE Conference on
  Computer Vision and Pattern Recognition}, pp. \bibinfo{pages}{136--145}.
\bibitem[{Wang et~al.(2018a)Wang, Zhang, Wang, Lu, Yang, Ruan and
  Borji}]{wang2018detect}
\bibinfo{author}{Wang, T.}, \bibinfo{author}{Zhang, L.}, \bibinfo{author}{Wang,
  S.}, \bibinfo{author}{Lu, H.}, \bibinfo{author}{Yang, G.},
  \bibinfo{author}{Ruan, X.}, \bibinfo{author}{Borji, A.},
  \bibinfo{year}{2018}a.
\newblock \bibinfo{title}{Detect globally, refine locally: A novel approach to
  saliency detection}, in: \bibinfo{booktitle}{Proceedings of the IEEE
  Conference on Computer Vision and Pattern Recognition}, pp.
  \bibinfo{pages}{3127--3135}.
\bibitem[{Wang et~al.(2019)Wang, Chao, Weinberger and van~der
  Maaten}]{wang2019revisiting}
\bibinfo{author}{Wang, Y.}, \bibinfo{author}{Chao, W.},
  \bibinfo{author}{Weinberger, K.}, \bibinfo{author}{van~der Maaten, L.S.},
  \bibinfo{year}{2019}.
\newblock \bibinfo{title}{Revisiting nearest neighbor classification for
  few-shot learning}.
\newblock \bibinfo{journal}{Preprint} .
\bibitem[{Wang et~al.(2020)Wang, Xu, Liu, Zhang and Fu}]{wang2020instance}
\bibinfo{author}{Wang, Y.}, \bibinfo{author}{Xu, C.}, \bibinfo{author}{Liu,
  C.}, \bibinfo{author}{Zhang, L.}, \bibinfo{author}{Fu, Y.},
  \bibinfo{year}{2020}.
\newblock \bibinfo{title}{Instance credibility inference for few-shot
  learning}, in: \bibinfo{booktitle}{Proceedings of the IEEE/CVF Conference on
  Computer Vision and Pattern Recognition}, pp. \bibinfo{pages}{12836--12845}.
\bibitem[{Wang et~al.(2018b)Wang, Girshick, Hebert and Hariharan}]{wang2018low}
\bibinfo{author}{Wang, Y.X.}, \bibinfo{author}{Girshick, R.},
  \bibinfo{author}{Hebert, M.}, \bibinfo{author}{Hariharan, B.},
  \bibinfo{year}{2018}b.
\newblock \bibinfo{title}{Low-shot learning from imaginary data}, in:
  \bibinfo{booktitle}{Proceedings of the IEEE conference on computer vision and
  pattern recognition}, pp. \bibinfo{pages}{7278--7286}.
\bibitem[{Wei et~al.(2017)Wei, Luo, Wu and Zhou}]{wei2017selective}
\bibinfo{author}{Wei, X.S.}, \bibinfo{author}{Luo, J.H.}, \bibinfo{author}{Wu,
  J.}, \bibinfo{author}{Zhou, Z.H.}, \bibinfo{year}{2017}.
\newblock \bibinfo{title}{Selective convolutional descriptor aggregation for
  fine-grained image retrieval}.
\newblock \bibinfo{journal}{IEEE Transactions on Image Processing}
  \bibinfo{volume}{26}, \bibinfo{pages}{2868--2881}.
\bibitem[{Wei et~al.(2019a)Wei, Wang, Liu, Shen and Wu}]{wei2019piecewise}
\bibinfo{author}{Wei, X.S.}, \bibinfo{author}{Wang, P.}, \bibinfo{author}{Liu,
  L.}, \bibinfo{author}{Shen, C.}, \bibinfo{author}{Wu, J.},
  \bibinfo{year}{2019}a.
\newblock \bibinfo{title}{Piecewise classifier mappings: Learning fine-grained
  learners for novel categories with few examples}.
\newblock \bibinfo{journal}{IEEE Transactions on Image Processing}
  \bibinfo{volume}{28}, \bibinfo{pages}{6116--6125}.
\bibitem[{Wei et~al.(2019b)Wei, Wu and Cui}]{wei2019deep}
\bibinfo{author}{Wei, X.S.}, \bibinfo{author}{Wu, J.}, \bibinfo{author}{Cui,
  Q.}, \bibinfo{year}{2019}b.
\newblock \bibinfo{title}{Deep learning for fine-grained image analysis: A
  survey}.
\newblock \bibinfo{journal}{arXiv preprint arXiv:1907.03069} .
\bibitem[{Wei et~al.(2018)Wei, Xie, Wu and Shen}]{wei2018mask}
\bibinfo{author}{Wei, X.S.}, \bibinfo{author}{Xie, C.W.}, \bibinfo{author}{Wu,
  J.}, \bibinfo{author}{Shen, C.}, \bibinfo{year}{2018}.
\newblock \bibinfo{title}{Mask-cnn: Localizing parts and selecting descriptors
  for fine-grained bird species categorization}.
\newblock \bibinfo{journal}{Pattern Recognition} \bibinfo{volume}{76},
  \bibinfo{pages}{704--714}.
\bibitem[{Xiao et~al.(2015)Xiao, Xu, Yang, Zhang, Peng and
  Zhang}]{xiao2015application}
\bibinfo{author}{Xiao, T.}, \bibinfo{author}{Xu, Y.}, \bibinfo{author}{Yang,
  K.}, \bibinfo{author}{Zhang, J.}, \bibinfo{author}{Peng, Y.},
  \bibinfo{author}{Zhang, Z.}, \bibinfo{year}{2015}.
\newblock \bibinfo{title}{The application of two-level attention models in deep
  convolutional neural network for fine-grained image classification}, in:
  \bibinfo{booktitle}{Proceedings of the IEEE Conference on Computer Vision and
  Pattern Recognition}, pp. \bibinfo{pages}{842--850}.
\bibitem[{Zhang et~al.(2016a)Zhang, Xu, Elhoseiny, Huang, Zhang, Elgammal and
  Metaxas}]{zhang2016spda}
\bibinfo{author}{Zhang, H.}, \bibinfo{author}{Xu, T.},
  \bibinfo{author}{Elhoseiny, M.}, \bibinfo{author}{Huang, X.},
  \bibinfo{author}{Zhang, S.}, \bibinfo{author}{Elgammal, A.},
  \bibinfo{author}{Metaxas, D.}, \bibinfo{year}{2016}a.
\newblock \bibinfo{title}{Spda-cnn: Unifying semantic part detection and
  abstraction for fine-grained recognition}, in:
  \bibinfo{booktitle}{Proceedings of the IEEE Conference on Computer Vision and
  Pattern Recognition}, pp. \bibinfo{pages}{1143--1152}.
\bibitem[{Zhang et~al.(2014)Zhang, Donahue, Girshick and
  Darrell}]{zhang2014part}
\bibinfo{author}{Zhang, N.}, \bibinfo{author}{Donahue, J.},
  \bibinfo{author}{Girshick, R.}, \bibinfo{author}{Darrell, T.},
  \bibinfo{year}{2014}.
\newblock \bibinfo{title}{Part-based r-cnns for fine-grained category
  detection}, in: \bibinfo{booktitle}{European Conference on Computer Vision},
  \bibinfo{organization}{Springer}. pp. \bibinfo{pages}{834--849}.
\bibitem[{Zhang et~al.(2018)Zhang, Che, Ghahramani, Bengio and
  Song}]{zhang2018metagan}
\bibinfo{author}{Zhang, R.}, \bibinfo{author}{Che, T.},
  \bibinfo{author}{Ghahramani, Z.}, \bibinfo{author}{Bengio, Y.},
  \bibinfo{author}{Song, Y.}, \bibinfo{year}{2018}.
\newblock \bibinfo{title}{Metagan: An adversarial approach to few-shot
  learning}, in: \bibinfo{booktitle}{Advances in Neural Information Processing
  Systems}, pp. \bibinfo{pages}{2365--2374}.
\bibitem[{Zhang et~al.(2019)Zhang, Jia and Wang}]{zhang2019part}
\bibinfo{author}{Zhang, Y.}, \bibinfo{author}{Jia, K.}, \bibinfo{author}{Wang,
  Z.}, \bibinfo{year}{2019}.
\newblock \bibinfo{title}{Part-aware fine-grained object categorization using
  weakly supervised part detection network}.
\newblock \bibinfo{journal}{IEEE Transactions on Multimedia} .
\bibitem[{Zhang et~al.(2015)Zhang, Wei, Wu, Cai, Lu, Nguyen and
  Do}]{zhang2015weakly}
\bibinfo{author}{Zhang, Y.}, \bibinfo{author}{Wei, X.s.}, \bibinfo{author}{Wu,
  J.}, \bibinfo{author}{Cai, J.}, \bibinfo{author}{Lu, J.},
  \bibinfo{author}{Nguyen, V.A.}, \bibinfo{author}{Do, M.N.},
  \bibinfo{year}{2015}.
\newblock \bibinfo{title}{Weakly supervised fine-grained image categorization}.
\newblock \bibinfo{journal}{arXiv preprint arXiv:1504.04943} .
\bibitem[{Zhang et~al.(2016b)Zhang, Wei, Wu, Cai, Lu, Nguyen and
  Do}]{zhang2016weakly}
\bibinfo{author}{Zhang, Y.}, \bibinfo{author}{Wei, X.S.}, \bibinfo{author}{Wu,
  J.}, \bibinfo{author}{Cai, J.}, \bibinfo{author}{Lu, J.},
  \bibinfo{author}{Nguyen, V.A.}, \bibinfo{author}{Do, M.N.},
  \bibinfo{year}{2016}b.
\newblock \bibinfo{title}{Weakly supervised fine-grained categorization with
  part-based image representation}.
\newblock \bibinfo{journal}{IEEE Transactions on Image Processing}
  \bibinfo{volume}{25}, \bibinfo{pages}{1713--1725}.
\bibitem[{Zhao et~al.(2017)Zhao, Wu, Feng, Peng and Yan}]{zhao2017diversified}
\bibinfo{author}{Zhao, B.}, \bibinfo{author}{Wu, X.}, \bibinfo{author}{Feng,
  J.}, \bibinfo{author}{Peng, Q.}, \bibinfo{author}{Yan, S.},
  \bibinfo{year}{2017}.
\newblock \bibinfo{title}{Diversified visual attention networks for
  fine-grained object classification}.
\newblock \bibinfo{journal}{IEEE Transactions on Multimedia}
  \bibinfo{volume}{19}, \bibinfo{pages}{1245--1256}.
\bibitem[{Zhao et~al.(2015)Zhao, Ouyang, Li and Wang}]{zhao2015saliency}
\bibinfo{author}{Zhao, R.}, \bibinfo{author}{Ouyang, W.}, \bibinfo{author}{Li,
  H.}, \bibinfo{author}{Wang, X.}, \bibinfo{year}{2015}.
\newblock \bibinfo{title}{Saliency detection by multi-context deep learning},
  in: \bibinfo{booktitle}{Proceedings of the IEEE Conference on Computer Vision
  and Pattern Recognition}, pp. \bibinfo{pages}{1265--1274}.
\bibitem[{Zheng et~al.(2017)Zheng, Fu, Mei and Luo}]{zheng2017learning}
\bibinfo{author}{Zheng, H.}, \bibinfo{author}{Fu, J.}, \bibinfo{author}{Mei,
  T.}, \bibinfo{author}{Luo, J.}, \bibinfo{year}{2017}.
\newblock \bibinfo{title}{Learning multi-attention convolutional neural network
  for fine-grained image recognition}, in: \bibinfo{booktitle}{Proceedings of
  the IEEE International Conference on Computer Vision}, pp.
  \bibinfo{pages}{5209--5217}.
\bibitem[{Zhong et~al.(2018)Zhong, Jiang, Zhang, Ji and Xiong}]{zhong2018multi}
\bibinfo{author}{Zhong, W.}, \bibinfo{author}{Jiang, L.},
  \bibinfo{author}{Zhang, T.}, \bibinfo{author}{Ji, J.},
  \bibinfo{author}{Xiong, H.}, \bibinfo{year}{2018}.
\newblock \bibinfo{title}{A multi-part convolutional attention network for
  fine-grained image recognition}, in: \bibinfo{booktitle}{2018 24th
  International Conference on Pattern Recognition (ICPR)},
  \bibinfo{organization}{IEEE}. pp. \bibinfo{pages}{1857--1862}.
\bibitem[{Ziko et~al.(2020)Ziko, Dolz, Granger and Ayed}]{ziko2020laplacian}
\bibinfo{author}{Ziko, I.}, \bibinfo{author}{Dolz, J.},
  \bibinfo{author}{Granger, E.}, \bibinfo{author}{Ayed, I.B.},
  \bibinfo{year}{2020}.
\newblock \bibinfo{title}{Laplacian regularized few-shot learning}, in:
  \bibinfo{booktitle}{International Conference on Machine Learning},
  \bibinfo{organization}{PMLR}. pp. \bibinfo{pages}{11660--11670}.

\end{thebibliography}

\end{document}